\documentclass[journal]{IEEEtran}
    \usepackage{blindtext}
    \usepackage{graphicx}
    \usepackage{mathrsfs}
    \usepackage{mathtools}
    \usepackage{booktabs}
    \usepackage{amsmath,amssymb}
    \usepackage{algorithm}
    \usepackage[noend]{algpseudocode}
    \usepackage{placeins}
    \usepackage{afterpage}

    \usepackage[dvipsnames]{xcolor}

\begin{document}                                              
    \ifCLASSINFOpdf
    \else
    \fi
    
    \hyphenation{op-tical net-works semi-conduc-tor}

    \bstctlcite{IEEEexample:BSTcontrol}
    %
    \title{Deep Learning for Electromyographic Hand Gesture Signal Classification Using Transfer Learning}
    
    \author{Ulysse~C\^ot\'e-Allard,
            Cheikh Latyr Fall,
            Alexandre Drouin,
    
            Alexandre Campeau-Lecours,        
            Cl\'ement Gosselin,
            Kyrre Glette,
            Fran\c{c}ois Laviolette$\dagger$,
            and~Benoit~Gosselin$\dagger$}
    
    \renewcommand\footnotemark{}
    \renewcommand\footnoterule{}
    \thanks{Ulysse C\^ot\'e-Allard*, Cheikh Latyr Fall and Benoit Gosselin are with the Department of Computer and Electrical Engineering, Alexandre Drouin and Fran\c{c}ois Laviolette are with the Department of Computer Science and Software Engineering, Alexandre Campeau-Lecours and Cl\'ement Gosselin are with the Department of Mechanical Engineering, Universit\'e Laval, Qu\'ebec, Qu\'ebec, Canada. Kyrre Glette is with RITMO and the Department of Informatics, University of Oslo, Oslo, Norway.  
    
    *Contact author email: ulysse.cote-allard.1@ulaval.ca

    $\dagger$ These authors share senior authorship.
    }
    \maketitle

    \begin{abstract}
    In recent years, deep learning algorithms have become increasingly more prominent for their unparalleled ability to automatically learn discriminant features from large amounts of data. However, within the field of electromyography-based gesture recognition, deep learning algorithms are seldom employed as they require an unreasonable amount of effort from a single person, to generate tens of thousands of examples.
            
    This work's hypothesis is that general, informative features can be learned from the large amounts of data generated by aggregating the signals of multiple users, thus reducing the recording burden while enhancing gesture recognition. Consequently, this paper proposes applying transfer learning on aggregated data from multiple users, while leveraging the capacity of deep learning algorithms to learn discriminant features from large datasets. Two datasets comprised of 19 and 17 able-bodied participants respectively (the first one is employed for pre-training) were recorded for this work, using the Myo Armband. A third Myo Armband dataset was taken from the NinaPro database and is comprised of 10 able-bodied participants. Three different deep learning networks employing three different modalities as input (raw EMG, Spectrograms and Continuous Wavelet Transform (CWT)) are tested on the second and third dataset. The proposed transfer learning scheme is shown to systematically and significantly enhance the performance for all three networks on the two datasets, achieving an offline accuracy of 98.31\% for 7 gestures over 17 participants for the CWT-based ConvNet and 68.98\% for 18 gestures over 10 participants for the raw EMG-based ConvNet.  Finally, a use-case study employing eight
    able-bodied participants suggests that real-time feedback allows users to adapt their muscle activation strategy which reduces the degradation in accuracy normally experienced over time.

    \end{abstract}
    
    
    \IEEEpeerreviewmaketitle

    \section{Introduction}
    
    Robotics and artificial intelligence can be leveraged to increase the autonomy of people living with disabilities. This is accomplished, in part, by enabling users to seamlessly interact with robots to complete their daily tasks with increased independence. In the context of hand prosthetic control, muscle activity provides an intuitive interface on which to perform hand gesture recognition~\cite{emg_gesture_classification_survey}. This activity can be recorded by surface electromyography (sEMG), a non-invasive technique widely adopted both in research and clinical settings. The sEMG signals, which are non-stationary, represent the sum of subcutaneous motor action potentials generated through muscular contraction~\cite{emg_gesture_classification_survey}. Artificial intelligence can then be leveraged as the bridge between sEMG signals and the prosthetic behavior. 
    
    The literature on sEMG-based gesture recognition primarily focuses on feature engineering, with the goal of characterizing sEMG signals in a discriminative way~\cite{emg_gesture_classification_survey, list_features, complex_features_explanation}. Recently, researchers have proposed deep learning approaches~\cite{allard2016convolutional, CNN_NinaPro_60_percent, cnn_da_inter_session}, shifting the paradigm from feature engineering to feature learning. Regardless of the method employed, the end-goal remains the improvement of the classifier's robustness. One of the main factors for accurate predictions, especially when working with deep learning algorithms, is the amount of training data available. Hand gesture recognition creates a peculiar context where a single user cannot realistically be expected to generate tens of thousands of examples in a single sitting. Large amounts of data can however be obtained by aggregating the recordings of multiple participants, thus fostering the conditions necessary to learn a general mapping of users' sEMG signal. This mapping might then facilitate the hand gestures' discrimination task with new subjects. Consequently, deep learning offers a particularly attractive context from which to develop a Transfer Learning (TL) algorithm to leverage inter-user data by pre-training a model on multiple subjects before training it on a new participant.
    
    As such, the main contribution of this work is to present a new TL scheme employing a convolutional network (ConvNet) to leverage inter-user data within the context of sEMG-based gesture recognition. A previous work~\cite{smc_transfer_learning} has already shown that learning simultaneously from multiple subjects significantly enhances the ConvNet's performance whilst reducing the size of the required training dataset typically seen with deep learning algorithms. This paper expands upon the aforementioned conference paper's work, improving the TL algorithm to reduce its computational load and improving its performance. Additionally, three new ConvNet architectures, employing three different input modalities, specifically designed for the robust and efficient classification of sEMG signals are presented. The raw signal, short-time Fourier transform-based spectrogram and Continuous Wavelet Transform (CWT) are considered for the characterization of the sEMG signals to be fed to these ConvNets. To the best of the authors' knowledge, this is the first time that CWTs are employed as features for the classification of sEMG-based hand gesture recognition (although they have been proposed for the analysis of myoelectric signals~\cite{CWT_emg_analyze}). Another major contribution of this article is the publication of a new sEMG-based gesture classification dataset comprised of 36 able-bodied participants. This dataset and the implementation of the ConvNets along with their TL augmented version are made readily available\footnote{https://github.com/Giguelingueling/MyoArmbandDataset}. Finally, this paper further expands the aforementioned conference paper by proposing a use-case experiment on the effect of real-time feedback on the online performance of a classifier without recalibration over a period of fourteen days. Note that, due to the stochastic nature of the algorithms presented in this paper, unless stated otherwise, all experiments are reported as an average of 20 runs. 
    
    This paper is organized as follows. An overview of the related work in hand gesture recognition through deep learning and transfer learning/domain adaptation is given in Sec.~\ref{related_work}. Sec.~\ref{dataset} presents the proposed new hand gesture recognition dataset, with data acquisition and processing details alongside an overview of the NinaPro DB5 dataset. A presentation of the different state-of-the-art feature sets employed in this work is given in Sec.~\ref{feature_extraction}. Sec.~\ref{deep_learning} thoroughly describes the proposed networks' architectures, while Sec.~\ref{transfer_learning} presents the TL algorithm used to augment said architecture. Moreover, comparisons with the state-of-the-art in gesture recognition are given in Sec.~\ref{comparison_classifiers}. A real-time use-case experiment on the ability of users to counteract signal drift from sEMG signals is presented in Sec.~\ref{medium_term_experiment}. Finally, results are discussed in Sec.~\ref{Discussion}.

    \section{Related Work}
    \label{related_work}
    
    sEMG signals can vary significantly between subjects, even when precisely controlling for electrode placement~\cite{emg_change_from_subject_to_subject}. Regardless, classifiers trained from a user can be applied to new participants achieving slightly better than random performances~\cite{emg_change_from_subject_to_subject} and high accuracy (85\% over 6 gestures) when augmented with TL on never before seen subjects~\cite{EMG_TL_on_never_seen_before_subject}. As such, sophisticated techniques have been proposed to leverage inter-user information. For example, research has been done to find a projection of the feature space that bridges the gap between an original subject and a new user~\cite{DA_same_feature_space_1, DA_same_feature_space_2}. Several works have also proposed leveraging a pre-trained model removing the need to simultaneously work with data from multiple users \cite{DA_model_knowledge_transfer, DA_multi_kernel_adaptive_learning, least_square_svm_domain_adaptation}. These non-deep learning TL approaches showed important performance gains compared to their non-augmented versions. Although, some of these gains might be due to the baseline's poorly optimized hyperparameters~\cite{da_is_useless}.

    Short-Time Fourier Transform (STFT) have been sparsely employed in the last decades for the classification of sEMG data~\cite{sEMG_stft_vs_wavelets, stft_emg_2006}. A possible reason for this limited interest in STFT is that much of the research on sEMG-based gesture recognition focuses on designing feature ensembles~\cite{list_features}. Because STFT on its own generates large amounts of features and are relatively computationally expensive, they can be challenging to integrate with other feature types. Additionally, STFTs have also been shown to be less accurate than Wavelet Transforms~\cite{sEMG_stft_vs_wavelets} on their own for the classification of sEMG data. Recently however, STFT features, in the form of spectrograms, have been applied as input feature space for the classification of sEMG data by leveraging ConvNets~\cite{allard2016convolutional, cnn_da_inter_session}.
    
    CWT features have been employed for electrocardiogram analysis~\cite{wavelet_ecg}, electroencephalography~\cite{wavelet_eeg} and EMG signal analysis, but mainly for lower limbs~\cite{quadriceps_cwt_emg, gait_cwt_emg}. Wavelet-based features have been used in the past for sEMG-based hand gesture recognition~\cite{wavelet_emg}. The features employed however, are based on the Discrete Wavelet Transform~\cite{DWT_sEMG_classification} and the Wavelet Packet Transform (WPT)~\cite{sEMG_stft_vs_wavelets} instead of the CWT. This preference might be due to the fact that both DWT and WPT are less computationally expensive than the CWT and are thus better suited to be integrated into an ensemble of features. Similarly to spectrograms however, CWT offers an attractive image-like representation to leverage ConvNets for sEMG signal classification and can now be efficiently implemented on embedded systems (see Appendix~\ref{embeddedSystems}). To the best of the authors' knowledge, this is the first time that CWT is utilized for sEMG-based hand gesture recognition.
    
    Recently, ConvNets have started to be employed for hand gesture recognition using single array~\cite{allard2016convolutional, CNN_NinaPro_60_percent} and matrix~\cite{CNN_matrix_emg} of electrodes. Additionally, other authors applied deep learning in conjunction with domain adaptation techniques~\cite{cnn_da_inter_session} but for inter-session classification as opposed to the inter-subject context of this paper. A thorough overview of deep learning techniques applied to EMG classification is given in~\cite{deepLearningEmgReview}. To the best of our knowledge, this paper, which is an extension of~\cite{smc_transfer_learning}, is the first time inter-user data is leveraged through TL for training deep learning algorithms on sEMG data.

    \section{sEMG datasets}
    \label{dataset}
    \subsection{Myo Dataset}
     \label{MyoDataset}
    One of the major contributions of this article is to provide a new, publicly available, sEMG-based hand gesture recognition dataset, referred to as the \textit{Myo Dataset}. This dataset contains two distinct sub-datasets with the first one serving as the \textit{pre-training dataset} and the second as the \textit{evaluation dataset}. The former, which is comprised of 19 able-bodied participants, should be employed to build, validate and optimize classification techniques. The latter, comprised of 17 able-bodied participants, is utilized only for the final testing. To the best of our knowledge, this is the largest dataset published utilizing the commercially available Myo Armband (Thalmic Labs) and it is our hope that it will become a useful tool for the sEMG-based hand gesture classification community.
    
    The data acquisition protocol was approved by the Comit\'es d'\'Ethique de la Recherche avec des \^etres humains de l'Universit\'e Laval (approbation number: 2017-026/21-02-2016) and informed consent was obtained from all participants. 
    
    \subsubsection{sEMG Recording Hardware}
    The electromyographic activity of each subject's forearm was recorded with the Myo Armband; an 8-channel, dry-electrode, low-sampling rate (200\textit{Hz}), low-cost consumer-grade sEMG armband.
    
    The Myo is non-intrusive, as the dry-electrodes allow users to simply slip the bracelet on without any preparation. Comparatively, gel-based electrodes require the shaving and washing of the skin to obtain optimal contact between the subject's skin and electrodes. Unfortunately, the convenience of the Myo Armband comes with limitations regarding the quality and quantity of the sEMG signals that are collected. Indeed, dry electrodes, such as the ones employed in the Myo, are less accurate and robust to motion artifact than gel-based ones~\cite{gel_vs_dry}. Additionally, while the recommended frequency range of sEMG signals is 5-500\textit{Hz}~\cite{standardEMG} requiring a sampling frequency greater or equal to 1000\textit{Hz}, the Myo Armband is limited to 200\textit{Hz}. This information loss was shown to significantly impact the ability of various classifiers to differentiate between hand gestures~\cite{lower_armband_frequency_negatively_affect_performances}. As such, robust and adequate classification techniques are needed to process the collected signals accurately.


    
    \subsubsection{Time-Window Length}
    For real-time control in a closed loop, input latency is an important factor to consider. A maximum latency of 300\textit{ms} was first recommended in~\cite{300ms}. Even though more recent studies suggest that the latency should optimally be kept between 100-250\textit{ms}~\cite{125ms, window_size_250}, the performance of the classifier should take priority over speed~\cite{125ms, performanceOverSpeed}. As is the case in~\cite{smc_transfer_learning}, a window size of 260\textit{ms} was selected to achieve a reasonable number of samples between each prediction due to the low frequency of the Myo. 
    
    \subsubsection{Labeled Data Acquisition Protocol}
    \label{dataset_recording_subsection}
\begin{figure}[!htbp]
    \centering
    \includegraphics[width=.8\linewidth]{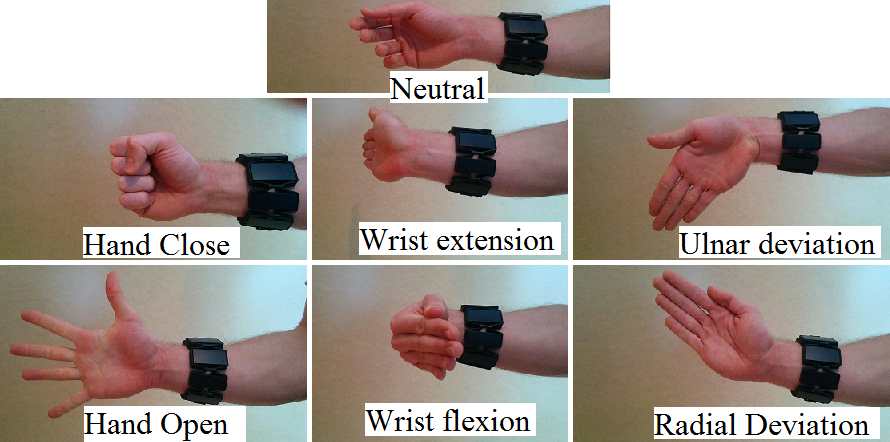}
    \caption{The 7 hand/wrist gestures considered in the \textit{Myo Dataset}.}
    \label{gestureImage}
    \end{figure}

    The seven hand/wrist gestures considered in this work are depicted in Fig.~\ref{gestureImage}. For both sub-datasets, the labeled data was created by requiring the user to hold each gesture for five seconds. The data recording was manually started by a researcher only once the participant correctly held the requested gesture. Generally, five seconds was given to the user between each gesture. This rest period was not recorded and as a result, the final dataset is balanced for all classes. The recording of the full seven gestures for five seconds is referred to as a \textit{cycle}, with four cycles forming a \textit{round}. In the case of the \textit{pre-training dataset}, a single \textit{round} is available per subject. For the \textit{evaluation dataset} three \textit{round}s are available with the first \textit{round} utilized for training (i.e. 140\textit{s} per participant) and the last two for testing (i.e. 240\textit{s} per participant). 
    
\begin{figure}[!htbp]
    \centering
    \includegraphics[width=.8\linewidth]{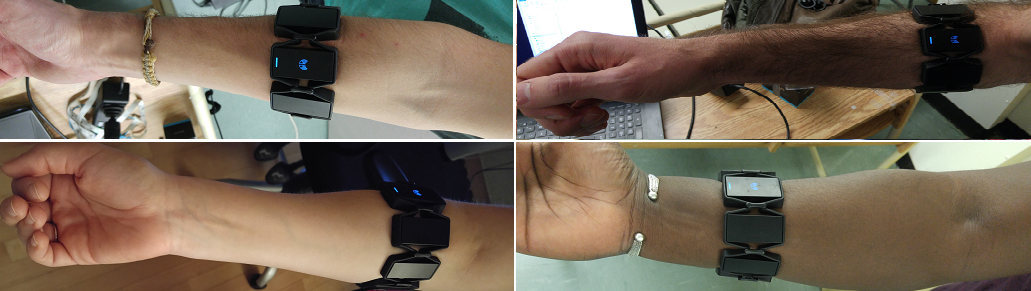}
    \caption{Examples of the range of armband placements on the subjects' forearm}
    \label{armbandPlacement}
    \end{figure}

    During recording, participants were instructed to stand up and have their forearm parallel to the floor and supported by themselves. For each of them, the armband was systematically tightened to its maximum and slid up the user's forearm, until the circumference of the armband matched that of the forearm. This was done in an effort to reduce bias from the researchers, and to emulate the wide variety of armband positions that end-users without prior knowledge of optimal electrode placement might use (see Fig.~\ref{armbandPlacement}). While the electrode placement was not controlled for, the orientation of the armband was always such that the blue light bar on the Myo was facing towards the hand of the subject. Note that this is the case for both left and right handed subjects. The raw sEMG data of the Myo is what is made available with this dataset. 
    
    Signal processing must be applied to efficiently train a classifier on the data recorded by the Myo armband. The data is first separated by applying sliding windows of 52 samples (260\textit{ms}) with an overlap of 235\textit{ms} (i.e. 7x190 samples for one cycle (5\textit{s} of data)). Employing windows of 260\textit{ms} allows 40\textit{ms} for the pre-processing and classification process, while still staying within the 300\textit{ms} target~\cite{300ms}. Note that utilizing sliding windows is viewed as a form of data augmentation in the present context (see Appendix~\ref{data_augmentation}). This is done for each gesture in each cycle on each of the eight channels. As such, in the dataset, an \textit{example} corresponds to the eight windows associated with their respective eight channels. From there, the processing depends on the classification techniques employed which will be detailed in Sec.~\ref{feature_extraction} and~\ref{deep_learning}.
    
    \subsection{NinaPro DB5}
    \label{ninaProDB5}
    The \textit{NinaPro DB5} is a dataset built to benchmark sEMG-based gesture recognition algorithms~\cite{NinaProDB5}. This dataset, which was recorded with the Myo Armband, contains data from 10 able-bodied participants performing a total of 53 different movements (including neutral) divided into three exercise sets. The second exercise set, which contains 17 gestures + neutral gesture, is of particular interest, as it includes all the gestures considered so far in this work. The 11 additional gestures which are presented in~\cite{NinaPro_nature} include wrist pronation, wrist supination and diverse finger extension amongst others. While this particular dataset was recorded with two Myo Armband, only the lower armband is considered as to allow direct comparison to the preceding dataset.
    
    \subsubsection{Data Acquisition and Processing}
    
    Each participant was asked to hold a gesture for five seconds followed by three seconds of neutral gesture and to repeat this action five more times (total of six repetitions). This procedure was repeated for all the movements contained within the dataset. The first four repetitions serve as the training set (20\textit{s} per gesture) and the last two (10\textit{s} per gesture) as the test set for each gesture. Note that the \textit{rest} movement (i.e. neutral gesture) was treated identically as the other gestures (i.e. first four repetitions for training (12\textit{s}) and the next two for testing (6\textit{s})).
    
    All data processing (e.g. window size, window overlap) are exactly as described in the previous sections.
    
    \section{Classic sEMG Classification}
    \label{feature_extraction}
    Traditionally, one of the most researched aspects of sEMG-based gesture recognition comes from feature engineering (i.e. manually finding a representation for sEMG signals that allows easy differentiation between gestures). Over the years, several efficient combinations of features both in the time and frequency domain have been proposed~\cite{sampen_pipeline, TD_stats, NinaPro, article_8}. This section presents the feature sets used in this work. See Appendix~\ref{feature_engineering} for a description of each feature. 
    
    \subsection{Feature Sets}
    \label{feature_sets}
    As this paper's main purpose is to present a deep learning-based TL approach to the problem of sEMG hand gesture recognition, contextualizing the performance of the proposed algorithms within the current state-of-the-art is essential. As such, four different feature sets were taken from the literature to serve as a comparison basis. The four feature sets will be tested on five of the most common classifiers employed for sEMG pattern recognition: Support Vector Machine (SVM)~\cite{NinaPro}, Artificial Neural Networks (ANN)~\cite{2009emg_review}, Random Forest (RF)~\cite{NinaPro}, K-Nearest Neighbors (KNN)~\cite{NinaPro} and Linear Discriminant Analysis (LDA)~\cite{article_8}. Hyperparameters for each classifier were selected by employing three fold cross-validation alongside random search, testing 50 different combinations of hyperparameters for each participant's dataset for each classifier. The hyperparameters considered for each classifier are presented in Appendix~\ref{hyperparameters_selection}.
    
    As is often the case, dimensionality reduction is applied~\cite{emg_gesture_classification_survey, complex_features_explanation, complexity_EMG}. LDA was chosen to perform feature projection as it is computationally inexpensive, devoid of hyperparameters and was shown to allow for robust classification accuracy for sEMG-based gesture recognition~\cite{article_8, lda_better_than_pca_for_EMG}. A comparison of the accuracy obtained with and without dimensionality reduction on the \textit{Myo Dataset} is given in Appendix~\ref{Dimensionality_reduction}. This comparison shows that in the vast majority of cases, the dimensionality reduction both reduced the computational load and enhanced the average performances of the feature sets. 
    
    The implementation employed for all the classifiers comes from the scikit-learn (v.1.13.1) Python package~\cite{scikit-learn}. The four feature sets employed for comparison purposes are:

    \subsubsection{Time Domain Features (TD)~\cite{TD_stats}}
    This set of features, which is probably the most commonly employed in the literature~\cite{lower_armband_frequency_negatively_affect_performances}, often serves as the basis for bigger feature sets~\cite{emg_gesture_classification_survey, article_8, NinaProDB5}. As such, TD is particularly well suited to serve as a baseline comparison for new classification techniques. The four features are: Mean Absolute Value (MAV), Zero Crossing (ZC), Slope Sign Changes (SSC) and Waveform Length (WL). 
    
    \subsubsection{Enhanced TD~\cite{article_8}}
    This set of features includes the TD features in combination with Skewness, Root Mean Square (RMS), Integrated EMG (IEMG), Autoregression Coefficients (AR) (P=11) and the Hjorth Parameters. It was shown to achieve excellent performances on a setup similar to the one employed in this article.
    
    \subsubsection{Nina Pro Features~\cite{NinaPro, NinaProDB5}}
    This set of features was selected as it was found to perform the best in the article introducing the NinaPro dataset. The set consists of the 
    the following features: RMS, Marginal Discrete Wavelet Transform (mDWT) (wavelet=db7, S=3), EMG Histogram (HIST) (bins=20, threshold=3$\sigma$) and the TD features. 
    
    \subsubsection{SampEn Pipeline~\cite{sampen_pipeline}}
    This last feature combination was selected among fifty features that were evaluated and ranked to find the most discriminating ones. The SampEn feature was ranked first amongst all the others. The best multi-features set found was composed of: SampEn(m=2, r=0.2$\sigma$), Cepstral Coefficient (order=4), RMS and WL.

    \section{Deep Learning Classifiers Overview}
    \label{deep_learning}
    
    ConvNets tend to be computationally expensive and thus ill-suited for embedded systems, such as those required when guiding a prosthetic. However, in recent years, algorithmic improvements and new hardware architectures have allowed for complex networks to run on very low power systems (see Appendix~\ref{embeddedSystems}). As previously mentioned, the inherent limitations of sEMG-based gesture recognition force the proposed ConvNets to contend with a limited amount of data from any single individual. To address the over-fitting issue, Monte Carlo Dropout (MC Dropout)~\cite{MCdropout}, Batch Normalization (BN)~\cite{BN}, and early stopping are employed.
    
    
    \subsection{Batch Normalization}
    BN is a technique that accelerates training and provides some form of regularization with the aims of maintaining a standard distribution of hidden layer activation values throughout training~\cite{BN}. BN accomplishes this by normalizing the mean and variance of each dimension of a batch of examples. To achieve this, a linear transformation based on two learned parameters is applied to each dimension. This process is done independently for each layer of the network. Once training is completed, the whole dataset is fed through the network one last time to compute the final normalization parameters in a layer-wise fashion. At test time, these parameters are applied to normalize the layer activations. BN was shown to yield faster training times whilst allowing better generalization.
    
    \subsection{Proposed Convolutional Network Architectures}
    \label{CNN_architecture}
\begin{figure}[!htbp]
    \centering
    \includegraphics[width=.35\linewidth]{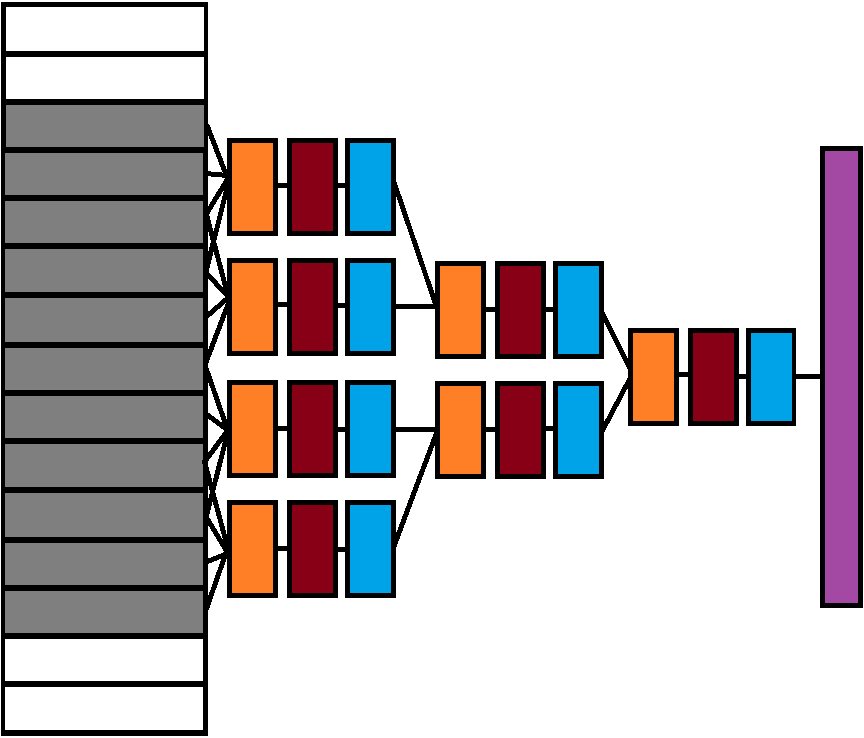}
    \caption{Typical slow-fusion ConvNet architecture~\cite{cnnVideo}. In this graph, the input (represented by grey rectangles) is a video (i.e. a sequence of images). The model separates the temporal part of the examples into disconnected parallel layers, which are then slowly fused together throughout the network.}
    \label{slow_fusion_normal}
    \end{figure}

    Videos are a representation of how spatial information (images) change through time. Previous works have combined this representation with ConvNets to address classification tasks~\cite{originalSlowFusion, cnnVideo}. One such successful algorithm is the slow-fusion model~\cite{cnnVideo} (see Fig.~\ref{slow_fusion_normal}).
    
    When calculating the spectrogram of a signal, the information is structured in a Time~x~Frequency fashion (Time~x~Scale for CWT). When the signal comes from an array of electrodes, these examples can naturally be structured as Time~x~Spatial~x~Frequency (Time~x~Spatial~x~Scale for CWT). As such, the motivation for using a slow-fusion architecture based ConvNet in this work is due to the similarities between videos data and the proposed characterization of sEMG signals, as both representations have analogous structures (i.e. Time~x~Spatial~x~Spatial for videos) and can describe non-stationary information. Additionally, the proposed architectures inspired by the slow-fusion model were by far the most successful of the ones tried on the pre-training dataset.

    
    
    \subsubsection{ConvNet for Spectrograms}
    
    The spectrograms, which are fed to the ConvNet, were calculated with Hann windows of length 28 and an overlap of 20 yielding a matrix of 4x15. The first frequency band was removed in an effort to reduce baseline drift and motion artifact. As the armband features eight channels, eight such spectrograms were calculated, yielding a final matrix of 4x8x14 (Time x Channel x Frequency).
    
\begin{figure*}[!htbp]
    \centering
    \includegraphics[width=\textwidth]{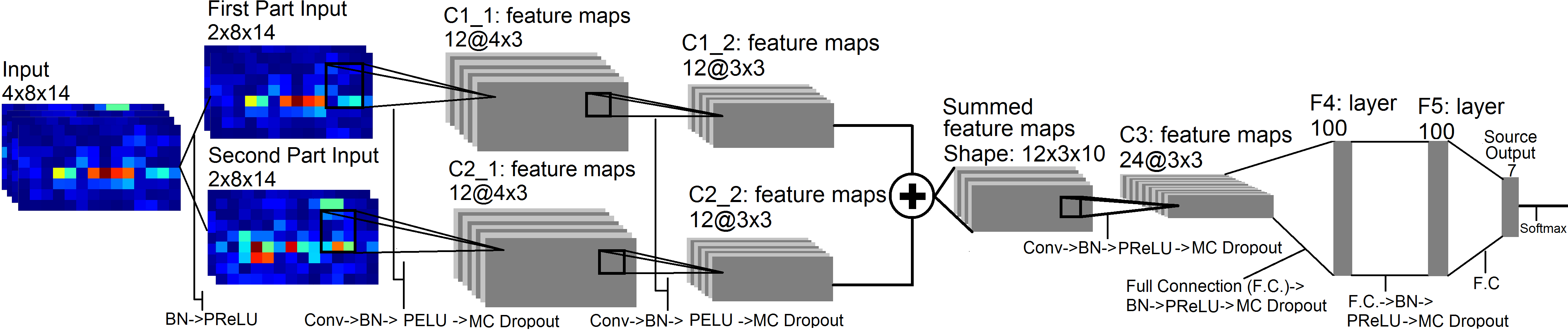}
    \caption{The proposed spectrogram ConvNet architecture to leverage spectrogram examples employing 67 179 learnable parameters. To allow the slow fusion process, the input is first separated equally into two parts with respect to the time axis. The two branches are then fused together by element-wise summing the feature maps together. In this figure, \textit{Conv} refer to \textit{Convolution} and \textit{F.C.} to \textit{Fully Connected} layers.}
    \label{stft_cnn_architecture}
    \end{figure*}

    The implementation of the spectrogram ConvNet architecture (see 
    Fig.~\ref{stft_cnn_architecture}) was created with Theano~\cite{theano} and Lasagne~\cite{lasagne}. As usual in deep learning, the architecture was created in a trial and error process taking inspiration from previous architectures (primarily~\cite{allard2016convolutional, cnn_da_inter_session, cnnVideo, smc_transfer_learning}). The non-linear activation functions employed are the parametric exponential linear unit (PELU)~\cite{PELU} and PReLU~\cite{PReLU}. ADAM~\cite{adam} is utilized for the optimization of the ConvNet (learning rate=$0.00681292$). The deactivation rate for MC Dropout is set at $0.5$ and the batch size at 128. Finally, to further reduce overfitting, early stopping is employed by randomly removing 10\% of the data from the training and using it as a validation set at the beginning of the optimization process. Note that learning rate annealing is applied with a factor of $5$ when the validation loss stops improving. The training stops when two consecutive decays occurs with no network performance amelioration on the validation set. All hyperparameter values were found by a random search on the \textit{pre-training dataset}. 

    \subsubsection{ConvNet for Continuous Wavelet Transforms}
\begin{figure*}[!htbp]
    \centering
    \includegraphics[width=\textwidth]{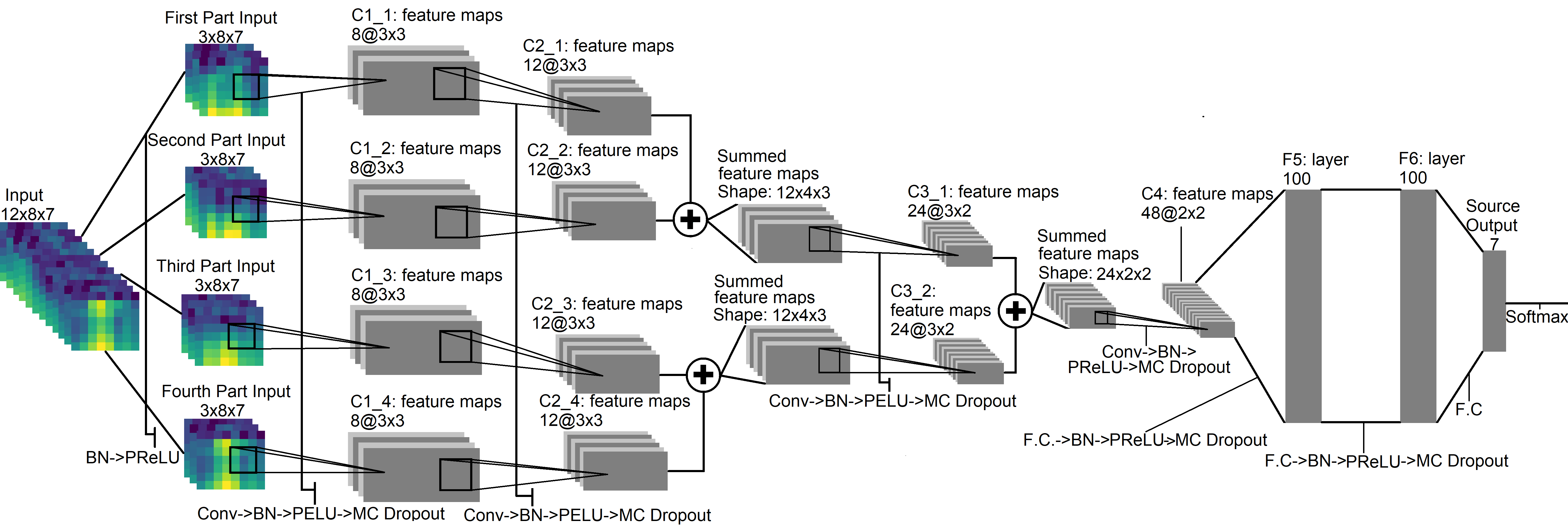}
    \caption{The proposed CWT ConvNet architecture to leverage CWT examples using 30 219 learnable parameters. To allow the slow fusion process, the input is first separated equally into four parts with respect to the time axis. The four branches are then slowly fused together by element-wise summing the feature maps together. In this figure, \textit{Conv} refers to \textit{Convolution} and \textit{F.C.} to \textit{Fully Connected} layers.}
    \label{cwt_cnn_architecture}
    \end{figure*}

    The architecture for the CWT ConvNet, (
    Fig.~\ref{cwt_cnn_architecture}), was built in a similar fashion as the spectrogram ConvNet one. Both the \textit{Morlet} and \textit{Mexican Hat} wavelet were considered for this work due to their previous application in EMG-related work~\cite{big_review_emg, cwt_tremor}. In the end, the Mexican Hat wavelet was selected, as it was the best performing during cross-validation on the \textit{pre-training dataset}. The CWTs were calculated with 32 scales yielding a 32x52 matrix. Downsampling is then applied at a factor of 0.25 employing spline interpolation of order 0 to reduce the computational load of the ConvNet during training and inference. Following downsampling, similarly to the spectrogram, the last row of the calculated CWT was removed as to reduce baseline drift and motion artifact. Additionally, the last column of the calculated CWT was also removed as to provide an even number of time-columns from which to perform the slow-fusion process. The final matrix shape is thus 12x8x7 (i.e. Time x Channel x Scale). The MC Dropout deactivation rate, batch size, optimization algorithm, and activation functions remained unchanged. The learning rate was set at $0.0879923$ (found by cross-validation).

    \subsubsection{ConvNet for raw EMG}
\begin{figure*}[!htbp]
    \centering
    \includegraphics[width=\textwidth]{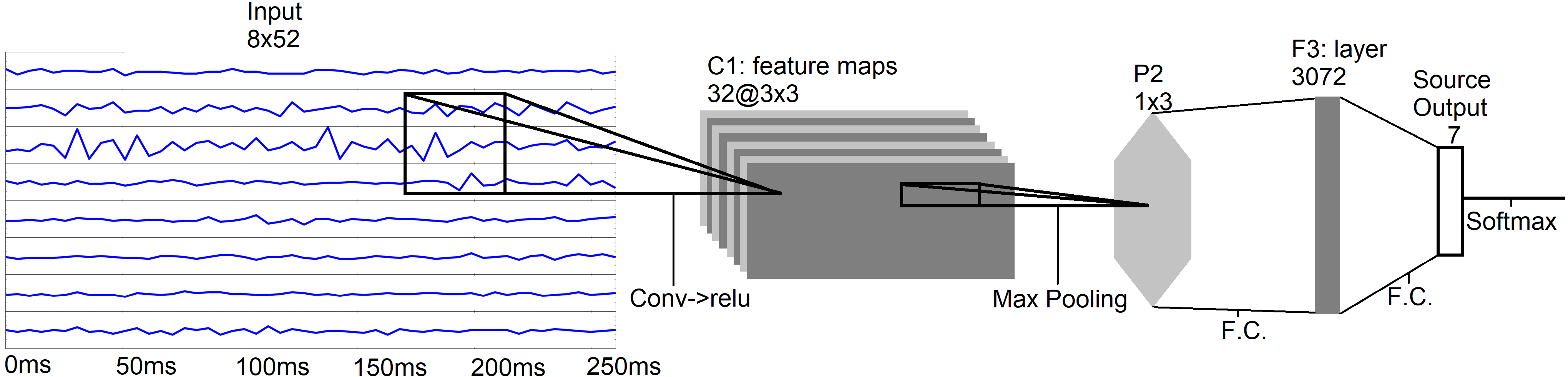}
    \caption{The raw ConvNet architecture to leverage raw EMG signals. In this figure, \textit{Conv} refers to \textit{Convolution} and \textit{F.C.} to \textit{Fully Connected} layers.}
    \label{raw_emg_architecture}
    \end{figure*}

\begin{figure*}[!htbp]
    \centering
    \includegraphics[width=\textwidth]{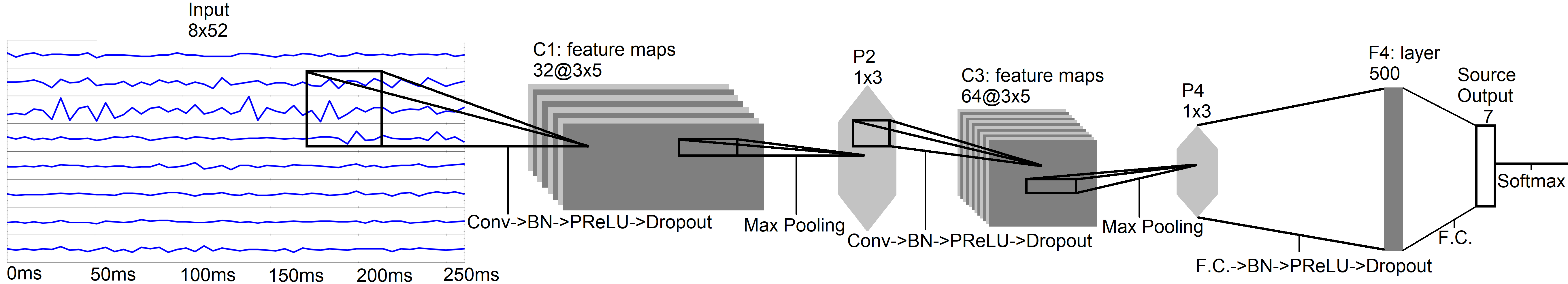}
    \caption{The enhanced raw ConvNet architecture using 549 091 learnable parameters. In this figure, \textit{Conv} refers to \textit{Convolution} and \textit{F.C.} to \textit{Fully Connected} layers.}
    \label{enhanced_raw_architecture}
    \end{figure*}

    A third ConvNet architecture taking the raw EMG signal as input is also considered. This network will help assess if employing time-frequency features lead to sufficient gains in accuracy performance to justify the increase in computational cost. As the raw EMG represents a completely different modality, a new type of architecture must be employed. To reduce bias from the authors as much as possible, the architecture considered is the one presented in~\cite{rawConvNetEMG}. The \textit{raw ConvNet} architecture can be seen in 
    Fig.~\ref{raw_emg_architecture}. This architecture was selected as it was also designed to classify a hand gesture dataset employing the Myo Armband. The architecture implementation (in PyTorch v.0.4.1) is exactly as described in~\cite{rawConvNetEMG} except for the learning rate (=$1.1288378916846883e-5$) which was found by cross-validation (tested 20 uniformly distributed values between $1e-6$ to $1e-1$ on a logarithm scale) and extending the length of the window size as to match with the rest of this manuscript. 
    The \textit{raw ConvNet} is further enhanced by introducing a second convolutional and pooling layer as well as adding dropout, BN, replacing RELU activation function with PReLU and using ADAM (learning rate=0.002335721469090121) as the optimizer. The \textit{enhanced raw ConvNet}'s architecture, which is shown in 
    Fig.~\ref{enhanced_raw_architecture}, achieves an average accuracy of 97.88\% compared to 94.85\% for the \textit{raw ConvNet}. Consequently, all experiments using raw emg as input will employ the \textit{raw enhanced ConvNet}.
    
    
    \section{Transfer Learning}
    \label{transfer_learning}
    One of the main advantages of deep learning comes from its ability to leverage large amounts of data for learning. As it would be too time-consuming for a single individual to record tens of thousands of examples, this work proposes to aggregate the data of multiple individuals. The main challenge thus becomes to find a way to leverage data from multiple users, with the objective of achieving higher accuracy with less data. TL techniques are well suited for such a task, allowing the ConvNets to generate more general and robust features that can be applied to a new subject's sEMG activity.
    
    As the data recording was purposefully as unconstrained as possible, the armband's orientation from one subject to another can vary widely. As such, to allow for the use of TL, automatic \textit{alignment} is a necessary first step. The alignment for each subject was made by identifying the most active channel (calculated using the IEMG feature) for each gesture on the first subject. On subsequent subjects, the channels were then circularly shifted until their activation for each gesture matched those of the first subject as closely as possible. 
    
    \subsection{Progressive Neural Networks}
    
    Fine-tuning is the most prevalent TL technique in deep learning~\cite{finetuning1, finetuning2}. It consists of training a model on a \textit{source domain} (abundance of labeled data) and using the trained weights as a starting point when presented with a new task. However, fine-tuning can suffer from \textit{catastrophic forgetting}~\cite{pnn}, where relevant and important features learned during pre-training are lost on the \textit{target domain} (i.e. new task). Moreover, by design, fine-tuning is ill-suited when significant differences exist between the source and the target, as it can bias the network into poorly adapted features for the task at hand. Progressive Neural Networks (PNN)~\cite{pnn} attempt to address these issues by pre-training a model on the source domain and freezing its weights. When a new task appears, a new network, with random initialization, is created and connected in a layer-wise fashion to the original network. This connection is done via non-linear lateral connections (See~\cite{pnn} for details). 
    
    \subsection{Adaptive Batch Normalization}
    
    In opposition to the PNN architecture, which uses a different network for the source and the target, AdaBatch employs the same network for both tasks. The TL occurs by freezing all the network's weights (learned during pre-training) when training on the target, except for the parameters associated with BN. The hypothesis behind this technique is that the label-related information (i.e. gestures) rests in the network model weights whereas the domain-related information (i.e. subjects) is stored in their BN statistic. In the present context, this idea can be generalized by applying a multi-stream AdaBatch scheme~\cite{cnn_da_inter_session}. Instead of employing one \textit{Source Network} per subject during pre-training, a single network is shared across all participants. However, the BN statistics from each subject are calculated independently from one another, allowing the ConvNet to extract more general and robust features across all participants. As such, when training the source network, the data from all subjects are aggregated and fed to the network together. It is important to note that each training batch is comprised solely of examples that belong to a single participant. This allows the update of the participant's corresponding BN statistic. 
    
    
    \subsection{Proposed Transfer Learning Architecture}
    \label{proposed_TL_architecture}
    
    The main tenet behind TL is that similar tasks can be completed in similar ways. The difficulty in this paper's context is then to learn a mapping between the source and target task as to leverage information learned during pre-training. Training one network per source-task (i.e. per participant) for the PNN is not scalable in the present context. However, by training a \textit{Source Network} (presented in Sec.~\ref{deep_learning}) shared across all participants of the \textit{pre-training dataset} with the multi-stream AdaBatch and adding only a second network for the target task using the PNN architecture, the scaling problem in the current context vanishes. This second network will hereafter be referred to as the \textit{Second Network}. The architecture of the \textit{Second Network} is almost identical to the \textit{Source Network}. The difference being in the activation functions employed. The \textit{Source Network} leveraged a combination of PReLU and PELU, whereas the \textit{Second Network} only employed PELU. This architecture choice was made through trial and error and cross-validation on the \textit{pre-training dataset}. Additionally, the weights of both networks are trained and initialized independently. During pre-training, only the \textit{Source Network} is trained to represent the information of all the participants in the \textit{pre-training dataset}. The parameters of the \textit{Source Network} are then frozen once pre-training is completed, except for the BN parameters as they represent the domain-related information and thus must retain the ability to adapt to new users.

\begin{figure}[!htbp]
    \centering
    \includegraphics[width=.8\linewidth]{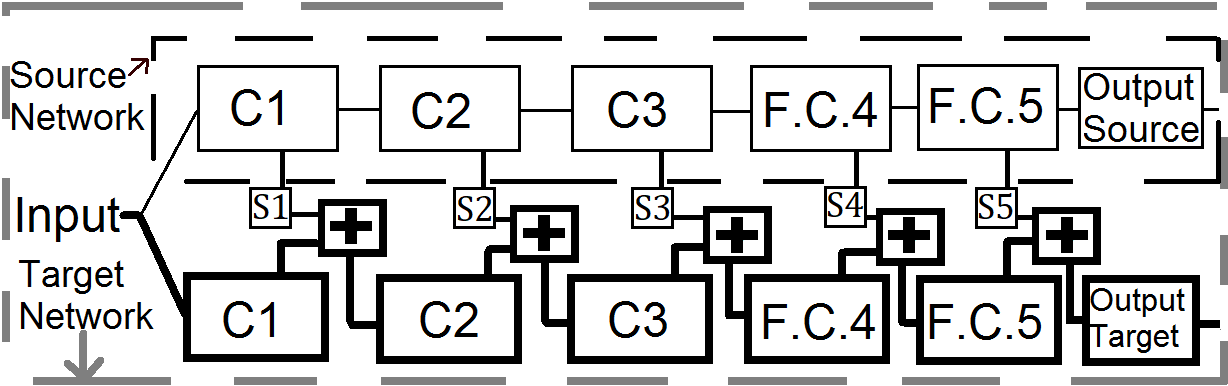}
    \caption{The PNN-inspired architecture. This figure represents the case with the spectrogram ConvNet. Note that the TL behavior is the same for the Raw-based or CWT-based ConvNet. C1,2,3 and F.C.4,5 correspond to the three stages of convolutions and two stages of fully connected layers respectively. The $Si$ (i=1..5) boxes represent a layer that scales its inputs by learned coefficients. The number of learned coefficients in one layer is the number of channels or the number of neurons for the convolutional and fully connected layers respectively. For clarity's sake, the slow fusion aspect is omitted from the representation although they are present for both the spectrogram and CWT-based ConvNet). The + boxes represent the merging through an element-wise summation of the ConvNets' corresponding layers.}
    \label{transfer_learning_architecture}
    \end{figure}

    Due to the application of the multi-stream AdaBatch scheme, the source task in the present context is to learn the \textit{general} mapping between muscle activity and gestures. One can see the problem of learning such mapping between the target and the source task as learning a residual of the source task. For this reason, the \textit{Source Network} shares information with the \textit{Second Network} through an element-wise summation in a layer-by-layer fashion (see Fig.~\ref{transfer_learning_architecture}). The idea behind the merging of information through element-wise summation is two-fold. First, compared to concatenating the features maps (as in \cite{smc_transfer_learning}) or employing non-linear lateral connections (like in~\cite{pnn}), element-wise summation minimizes the computational impact of connecting the \textit{Source Network} and the \textit{Second Network} together. Second, this provides a mechanism that fosters residual learning as inspired by Residual Networks~\cite{resnet}. Thus, the \textit{Second Network} only needs to learn weights that express the difference between the new target and source task. All outputs from the \textit{Source Network} layers to the \textit{Second Network} are multiplied by learnable coefficients before the sum-connection. This scalar layer provides an easy mechanism to neuter the \textit{Source Network's} influence on a layer-wise level. This is particularly useful if the new target task is so different that for some layers the information from the \textit{Source Network} actually hinders learning. Note that a single-stream scheme (i.e. all subjects share statistics and BN parameters are also frozen on the \textit{Source Network}) was also tried. As expected, this scheme's performances started to rapidly worsen as the number of source participants augmented, lending more credence to the initial AdaBatch hypothesis.
    
    The combination of the \textit{Source Network} and \textit{Second Network} will hereafter be referred to as the \textit{Target Network}. An overview of the final proposed architecture is presented in Fig.~\ref{transfer_learning_architecture}. During training of the \textit{Source Network} (i.e. pre-training), MC Dropout rate is set at 35\% and when training the \textit{Target Network} the rate is set at 50\%. Note that different architecture choices for the \textit{Source Network} and \textit{Second Network} were required to augment the performance of the system as a whole. This seems to indicate that the two tasks (i.e. learning a general mapping of hand gestures and learning a specific mapping), might be different enough that even greater differentiation through specialization of the two networks might increase the performance further. 
    
    \section{Classifier Comparison}
    \label{comparison_classifiers}

    \subsection{Myo Dataset}
    All pre-trainings in this section were done on the \textit{pre-training dataset} and all training (including for the traditional machine learning algorithms) were done on the first \textit{round} of the \textit{evaluation dataset}.
    \subsubsection{Comparison with Transfer Learning}
    \label{TL_MyoDataset}
    Considering each participant as a separate dataset allows for the application of the one-tail Wilcoxon signed-rank test~\cite{wilcoxon1945individual} ($n=17$). 
    Table~\ref{tableEvaluationDatasetTLComparison} shows a comparison of each ConvNet with their TL augmented version. Accuracies are given for one, two, three and four cycles of training.
    
    \begin{table*}[!htbp]
    \caption{Classification accuracy of the ConvNets on the \textit{Evaluation Dataset} with respect to the number of training cycles performed. }
    \centering
    \resizebox{\textwidth}{!}{
    \begin{tabular}{ccc|cc|cc}
    \hline
     & Raw & \begin{tabular}[c]{@{}c@{}}Raw + TL\end{tabular} & Spectrogram & \begin{tabular}[c]{@{}c@{}}Spectrogram + TL\end{tabular} & CWT & \begin{tabular}[c]{@{}c@{}}CWT + TL\end{tabular} \\ \hline
    \multicolumn{1}{c|}{4 Cycles} & 97.08\% & \textbf{97.39\%} & 97.14\% & \textbf{97.85\%} & 97.95\% & \textbf{98.31\%} \\
    \multicolumn{1}{c|}{STD} & 4.94\% & \textbf{4.07\%} & 2.85\% & \textbf{2.45\%} & 2.49\% & \textbf{2.16\%} \\
    \multicolumn{1}{c|}{H0 (p-value)} & 0 (0.02187) & \textbf{-} & 0 (0.00030) & - & 0 (0.00647) & - \\ \hline
    \multicolumn{1}{c|}{3 Cycles} & 96.22\% & \textbf{96.95\%} & 96.33\% & \textbf{97.40\%} & 97.22\% & \textbf{97.82\%} \\
    \multicolumn{1}{c|}{STD} & 6.49\% & \textbf{4.88\%} & 3.49\% & \textbf{2.91\%} & 3.46\% & \textbf{2.41\%} \\
    \multicolumn{1}{c|}{H0 (p-value)} & 0 (0.00155) & \textbf{-} & 0 (0.00018) & - & 0 (0.00113) & - \\ \hline
    \multicolumn{1}{c|}{2 Cycles} & 94.53\% & \textbf{95.49\%} & 94.19\% & \textbf{96.05\%} & 95.17\% & \textbf{96.63\%} \\
    \multicolumn{1}{c|}{STD} & 9.63\% & \textbf{7.26\%} & 5.95\% & \textbf{6.00\%} & 5.77\% & \textbf{4.54\%} \\
    \multicolumn{1}{c|}{H0 (p-value)} & 0 (0.00430) & \textbf{-} & 0 (0.00015) & - & 0 (0.00030) & - \\ \hline
    \multicolumn{1}{c|}{1 Cycle} & 89.04\% & \textbf{92.46\%} & 88.51\% & \textbf{93.93\%} & 89.02\% & \textbf{94.69\%} \\
    \multicolumn{1}{c|}{STD} & 10.63\% & \textbf{7.79\%} & 8.37\% & \textbf{6.56\%} & 10.24\% & \textbf{5.58\%} \\
    \multicolumn{1}{c|}{H0 (p-value)} & 0 (0.00018) & \textbf{-} & 0 (0.00015) & - & 0 (0.00015) & - \\ \hline
    \end{tabular}}

    * The \textit{one-tail Wilcoxon signed rank test} is applied to compare the ConvNet enhanced with the proposed TL algorithm to their non-augmented counterpart. Null hypothesis is rejected when $H_0=0$ ($p<0.05$).
    
    **The STD represents the pooled standard variation in accuracy for the 20 runs over the 17 participants. 
    \label{tableEvaluationDatasetTLComparison}
    \end{table*}
    \subsubsection{Comparison with State of the art}
    \label{comparison_deep_learning_MyoDataset}
    A comparison between the proposed CWT-based ConvNet and a variety of classifiers trained on the features sets presented in Sec.~\ref{feature_sets} is given in 
    Table~\ref{MyoDatasetResults}. 
    
    \begin{table*}[!htbp]
    \caption{Classifiers comparison on the \textit{Evaluation Dataset} with respect to the number of training cycles performed.}
    \label{MyoDatasetComparisonStateOfTHeArtTable}
    \centering
    \resizebox{\textwidth}{!}{
    \begin{tabular}{ccccccc}
    \hline
     & TD & Enhanced TD & Nina Pro & SampEn Pipeline & CWT & CWT + TL \\ \hline
    4 Cycles & 97.61\% (LDA) & 98.14\% (LDA) & 97.59\% (LDA) & 97.72\% (LDA) & 97.95\% & \textbf{98.31\%} \\
    STD & 2.63\% & 2.21\% & 2.74\% & 1.98\% & 2.49\% & \textbf{2.16\%} \\
    Friedman Rank & 3.94 & 2.71 & 4.29 & 3.47 & 3.94 & \textbf{2.65} \\
    H0 & 1 & 1 & 1 & 1 & 1 & \textbf{-} \\ \hline
    3 Cycles & 96.33\% (KNN) & 97.33\% (LDA) & 96.76\% (KNN) & 96.87\% (KNN) & 97.22\% & \textbf{97.82\%} \\
    STD & 6.11\% & 3.24\% & 3.85\% & 5.06\% & 3.46\% & \textbf{2.41\%} \\
    Friedman Rank & 4.41 & 2.77 & 4.05 & 3.53 & 3.94 & \textbf{2.29} \\
    H0 & 0 (0.00483) & 1 & 0 (0.02383) & 1 & 0 (0.03080) & \textbf{-} \\ \hline
    2 Cycles & 94.12\% (KNN) & 94.79\% (LDA) & 94.23\% (KNN) & 94.68\% (KNN) & 95.17\% & \textbf{96.63\%} \\
    STD & 9.08\% & 7.82\% & 7.49\% & 8.31\% & 5.77\% & \textbf{4.54\%} \\
    Friedman Rank & 4.41 & 3.24 & 4.41 & 3.29 & 3.65 & \textbf{2.00} \\
    H0 (adjusted p-value) & 0 (0.00085) & 1 & 0 (0.00085) & 1 & 0 (0.03080) & \textbf{-} \\ \hline
    1 Cycle & 90,77\% (KNN) & 91.25\% (LDA) & 90.21\% (LDA) & 91.66\% (KNN) & 89.02\% & \textbf{94.69\%} \\
    STD & 9.04\% & 9.44\% & 7.73\% & 8.74\% & 10.24\% & \textbf{5.58\%} \\
    Friedman Rank & 3.71 & 3.41 & 4.41 & 3.05 & 4.88 & \textbf{1.53} \\
    H0 (adjusted p-value) & 0 (0.00208) & 0 (0.00670) & 0 (0.00003) & 0 (0.01715) & 0 (\textless{}0.00001) & \textbf{-} \\ \hline
    \end{tabular}}
    \label{MyoDatasetResults}

    *For brevity's sake, only the best performing classifier for each feature set in each cycle is reported (indicated in parenthesis).

    **The STD represents the pooled standard variation in accuracy for the 20 runs over the 17 participants. 
    
    ***The Friedman Ranking Test followed by the Holm's post-hoc test is performed.
    \end{table*}
    
    As suggested in~\cite{FriedmanPlusHolm}, a two-step procedure is employed to compare the deep learning algorithms with the current state-of-the-art. First, Friedman's test ranks the algorithms amongst each other. Then, Holm's post-hoc test is applied ($n=17$) using the best ranked method as a comparison basis.

    

    \subsection{NinaPro Dataset}
    
    \subsubsection{Comparison with Transfer Learning}
    
    Performance of the proposed ConvNet architecture alongside their TL augmented versions are investigated on the \textit{NinaPro DB5}. As no specific pre-training dataset is available for the \textit{NinaPro DB5}, the pre-training for each participant is done employing the training sets of the remaining nine participants. 
    Table~\ref{NinaProTL_ConvNet_Table} shows the average accuracy over the 10 participants of the~\textit{NinaPro DB5} for one to four cycles. Similarly to Sec.~\ref{TL_MyoDataset}, the one-tail Wilcoxon Signed rank test is performed for each cycle between each ConvNet and their TL augmented version.
    
    
    \begin{table*}[!htbp]
    \caption{Classification accuracy of the ConvNets on the \textit{NinaPro DB5} with respect to the number of training cycles performed.}
    \label{NinaProTL_ConvNet_Table}
    \centering
    \resizebox{\textwidth}{!}{
    \begin{tabular}{ccc|cc|cc}
    \hline
     & Raw & \begin{tabular}[c]{@{}c@{}}Raw + TL\end{tabular} & Spectrogram & \begin{tabular}[c]{@{}c@{}}Spectrogram + TL\end{tabular} & CWT & \begin{tabular}[c]{@{}c@{}}CWT + TL\end{tabular} \\ \hline
    \multicolumn{1}{c|}{4 Repetitions} & 66.32\% & \textbf{68.98\%} & 63.60\% & \textbf{65.10\%} & 61.89\% & \textbf{65.57\%} \\
    \multicolumn{1}{c|}{STD} & 3.94\% & \textbf{4.46\%} & 3.94\% & \textbf{3.99\%} & 4.12\% & \textbf{3.68\%} \\
    \multicolumn{1}{c|}{H0 (p-value)} & 0 (0.00253) & \textbf{-} & 0 (0.00253) & - & 0 (0.00253) & - \\ \hline
    \multicolumn{1}{c|}{3 Repetitions} & 61.91\% & \textbf{65.16\%} & 60.09\% & \textbf{61.70\%} & 58.37\% & \textbf{62.21\%} \\
    \multicolumn{1}{c|}{STD} & 3.94\% & \textbf{4.46\%} & 4.03\% & \textbf{4.29\%} & 4.19\% & \textbf{3.93\%} \\
    \multicolumn{1}{c|}{H0 (p-value)} & 0 (0.00253) & \textbf{-} & 0 (0.00253) & - & 0 (0.00253) & - \\ \hline
    \multicolumn{1}{c|}{2 Repetitions} & 55.67\% & \textbf{60.12\%} & 55.35\% & \textbf{57.19\%} & 53.32\% & \textbf{57.53\%} \\
    \multicolumn{1}{c|}{STD} & 4.38\% & \textbf{4.79\%} & 4.50\% & \textbf{4.71\%} & 3.72\% & \textbf{3.69\%} \\
    \multicolumn{1}{c|}{H0 (p-value)} & 0 (0.00253) & \textbf{-} & 0 (0.00253) & - & 0 (0.00253) & - \\ \hline
    \multicolumn{1}{c|}{1 Repetitions} & 46.06\% & \textbf{49.41\%} & 45.59\% & \textbf{47.39\%} & 42.47\% & \textbf{48.33\%} \\
    \multicolumn{1}{c|}{STD} & 6.09\% & \textbf{5.82\%} & 5.58\% & \textbf{5.30\%} & 7.04\% & \textbf{5.07\%} \\
    \multicolumn{1}{c|}{H0 (p-value)} & 0 (0.00467) & \textbf{-} & 0 (0.00467) & - & 0 (0.00253) & - \\ \hline
    \end{tabular}}

    * The \textit{Wilcoxon signed rank test} is applied to compare the ConvNet enhanced with the proposed TL algorithm to their non-augmented counterpart. Null hypothesis is rejected when $H_0=0$ ($p<0.05$).
    
    **The STD represents the pooled standard variation in accuracy for the 20 runs over the 17 participants. 
    \end{table*}
    

    \subsubsection{Comparison with State of the art}
    \label{comparison_deep_learning_NinaPro}
    Similarly to Sec.~\ref{comparison_deep_learning_MyoDataset}, a comparison between the TL-augmented ConvNet and the traditional classifier trained on the state-of-the-art feature set is given in 
    Table~\ref{NinaProComparisonStateOfTheArt}. The accuracies are given for one, two, three and four cycles of training. A two-step statistical test with the Friedman test as the first step and Holm post-hoc as the second step is again employed. 
    
    
    \begin{table*}[!htbp]
    \caption{Classifiers Comparison on the \textit{NinaPro DB5} with respect to the number of repetitions used during training.}
    \centering
    \label{NinaProComparisonStateOfTheArt}
    \resizebox{\textwidth}{!}{
    \begin{tabular}{ccccccc}
    \hline
     & TD & Enhanced TD & Nina Pro & SampEn Pipeline & Raw & Raw + TL \\ \hline
    4 Repetitions & 59.91\% (RF) & 59.57\% (RF) & 56.72\% (RF) & 62.30\% (RF) & 66.32\% & \textbf{68.98\%} \\
    STD & 3.50\% & 4.43\% & 4.01\% & 3.94\% & 3.77\% & \textbf{4.09\%} \\
    Friedman Rank & 4.30 & 4.60 & 6.00 & 3.00 & 2.10 & \textbf{1.00} \\
    H0 (Adjusted p-value) & 0 (0.00024) & 0 (0.00007) & 0 (\textless 0.00001) & 0 (0.03365) & 1 & \textbf{-} \\ \hline
    3 Repetitions & 55.73\% (RF) & 55.32\% (RF) & 52.33\% (RF) & 58.24\% (RF) & 61.91\% & \textbf{65.16\%} \\
    STD & 3.75\% & 4.48\% & 4.63\% & 4.22\% & 3.94\% & \textbf{4.46\%} \\
    Friedman Rank & 4.40 & 4.60 & 6.00 & 3.00 & 2.00 & \textbf{1.00} \\
    H0 (Adjusted p-value) & 0 (0.00014) & 0 (0.00007) & 0 (\textless{}0.00001) & 0 (0.03365) & 1 & \textbf{-} \\ \hline
    2 Repetitions & 50.85\% (RF) & 50.08\% (LDA) & 46.85\% (LDA) & 53.00\% (RF) & 55.65\% & \textbf{60.12\%} \\
    STD & 4.29\% & 4.63\% & 4.81\% & 3.85\% & 4.38\% & \textbf{4.79\%} \\
    Friedman Rank & 4.20 & 4.60 & 6.00 & 3.10 & 2.10 & \textbf{1.00} \\
    H0 (Adjusted p-value) & 0 (0.00039) & 0 (0.00007) & 0 (\textless{}0.00001) & 0 (0.02415) & 1 & \textbf{-} \\ \hline
    1 Repetitions & 40.70\% (RF) & 40.86\% (LDA) & 37.60\% (LDA) & 42.26\% (LDA) & 46.06\% & \textbf{49.41\%} \\
    STD & 5.84\% & 6.91\% & 6.67\% & 5.78\% & 6.09\% & \textbf{5.82\%} \\
    Friedman Rank & 4.30 & 4.30 & 5.80 & 3.50 & 2.00 & \textbf{1.10} \\
    H0 (Adjusted p-value) & 0 (0.00052) & 0 (0.00052) & 0 (\textless{}0.00001) & 0 (0.00825) & 1 & \textbf{-} \\ \hline
    \end{tabular}}
    \label{NinaProResults}
    
    *For brevity's sake, only the best performing classifier for each feature set is reported (indicated in parenthesis).
    
    **The STD represents the pooled standard variation in accuracy for the 20 runs over the 17 participants.

    ***The Friedman Ranking Test followed by the Holm's post-hoc test is performed.
    \end{table*}


    \subsubsection{Out-of-Sample Gestures}

\begin{figure}[!htbp]
    \centering
    \includegraphics[width=.8\linewidth]{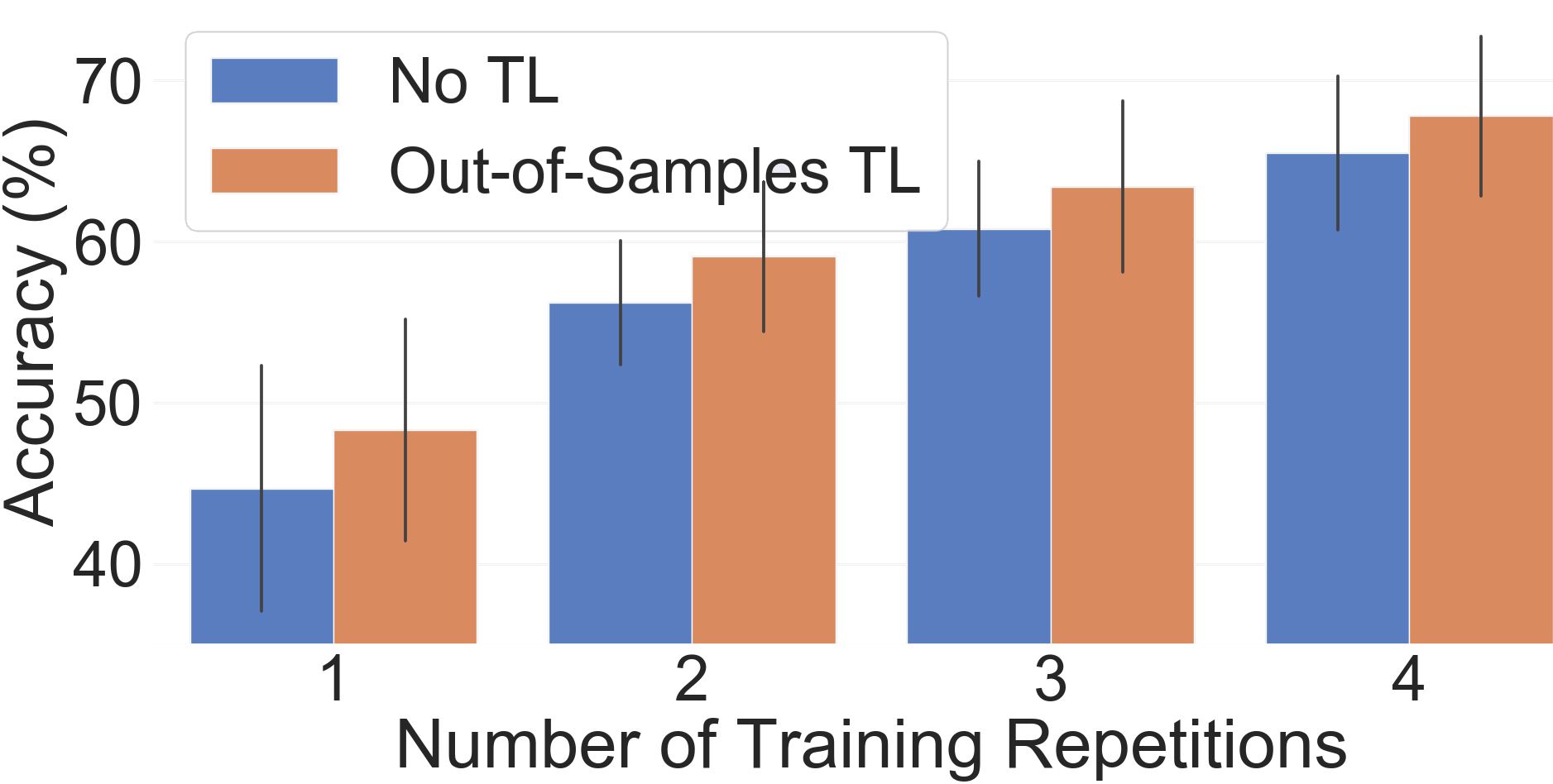}
    \caption{Classification accuracy of the CWT-based ConvNets on the \textit{NinaPro DB5} with respect to the number of repetitions employed during training. The pre-training was done using the \textit{pre-training dataset}. Training and testing only considered the 11 gestures from the \textit{NinaPro DB5} not included in the pre-training. The error bars correspond to the STD across all ten participants.}
    \label{NinaProAccuracyGroup}
    \end{figure}

    A final test involving the \textit{NinaPro DB5} was conducted to evaluate the impact on the proposed TL algorithm when the target is comprised solely of out-of-sample gestures (i.e. never-seen-before gestures). To do so, the proposed CWT ConvNet was trained and evaluated on the training and test set of the \textit{NinaPro DB5} as described before, but considering only the gestures that were absent from the \textit{pre-training dataset} (11 total). The CWT ConvNet was then compared to its TL augmented version which was pre-trained on the \textit{pre-training dataset}. Fig.~\ref{NinaProAccuracyGroup} presents the accuracies obtained for the classifiers with different number of repetitions employed for training. The difference in accuracy is considered statistically significant by the one-tail Wilcoxon Signed rank test for all cycles of training. Note that, similar, statistically significant results were obtained for the raw-based and spectrogram-based ConvNets.
    
    \section{Real-Time Classification and Medium Term Performances (case study)}
    \label{medium_term_experiment}
    This last experiment section proposes a use-case study of the online (i.e. real-time) performance of the classifier over a period of 14 days for eight able-bodied participants. In previous literature, it has been shown that, when no re-calibration occur, the performance of a classifier degrades over time due to the non-stationary property of sEMG signals~\cite{recalibration_necessity}. The main goal of this use-case experiment is to evaluate if users are able to self-adapt and improve the way they perform gestures based on visual feedback from complex classifiers (e.g. \textit{CWT+TL}), thus reducing the expected classification degradation.
    
    
    To achieve this, each participant recorded a training set as described in Sec.~\ref{dataset}. 
    Then, over the next fourteen days, a daily \textit{session} was recorded based on the participant's availability. A \textit{session} consisted of holding a set of 30 randomly selected gestures (among the seven shown in Fig.~\ref{gestureImage}) for ten seconds each, resulting in five minutes of continuous sEMG data. Note that to be more realistic, the participants began by placing the armband themselves, leading to slight armband position variations between sessions.
    
    The eight participants were randomly separated into two equal groups. The first group, referred to as the \textit{Feedback} group, received real-time feedback on the gesture predicted by the classifier in the form of text displayed on a computer screen. The second group, referred to as the \textit{Without Feedback} group, did not receive classifier feedback. The classifier employed in this experiment is the \textit{CWT+TL}, as it was the best performing classifier tested on the \textit{Evaluation Dataset}. Because the transitions are computer-specified, there is a latency between a new requested gesture and the participant's reaction. To reduce the impact of this phenomenon, the data from the first second after a new requested gesture is ignored from this section results. The number of data points generated by a single participant varies between 10 and 16 depending on the participant's availability during the experiment period.
    
\begin{figure}[!htbp]
    \centering
    \includegraphics[width=.9\linewidth]{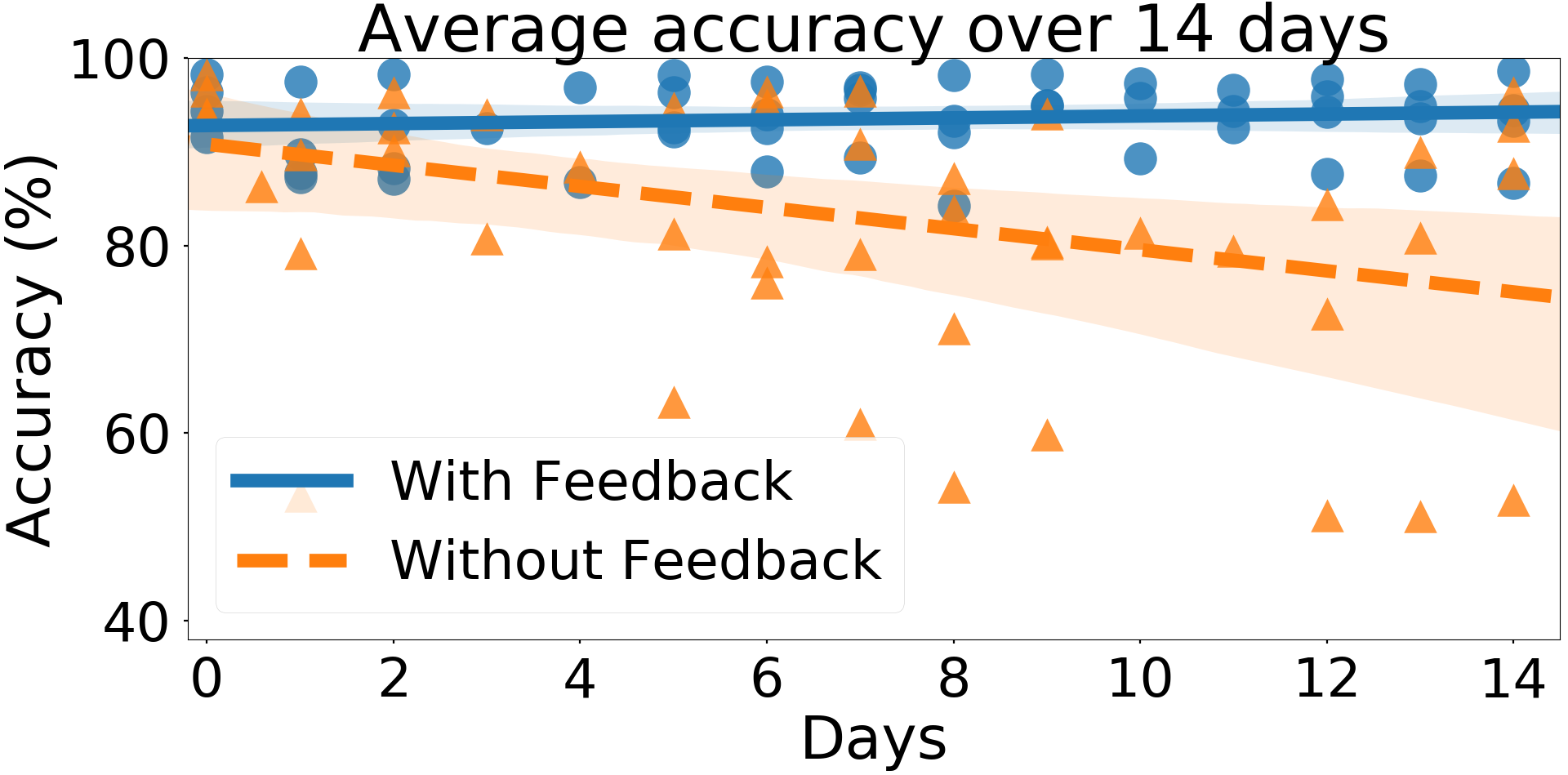}
    \caption{Average accuracy over 14 days without recalibration of the CWT+TL ConvNet
    The blue circles represent data from the \textit{Feedback} group whereas the orange triangles represent data from the \textit{Without Feedback} group.
    The translucent bands around the linear regressions represent the confidence interval (95\%) estimated by bootstrap.}
    \label{accuracy_14_days}
    \end{figure}

    As it can be observed in Fig.~\ref{accuracy_14_days}, while the \textit{Without Feedback} group did experience accuracy degradation over the 14 days, the \textit{Feedback} group was seemingly able to counteract this degradation. Note that, the average accuracy across all participants for the first recording session was 95.42\%.

\begin{figure}[!htbp]
    \centering
    \includegraphics[width=.9\linewidth]{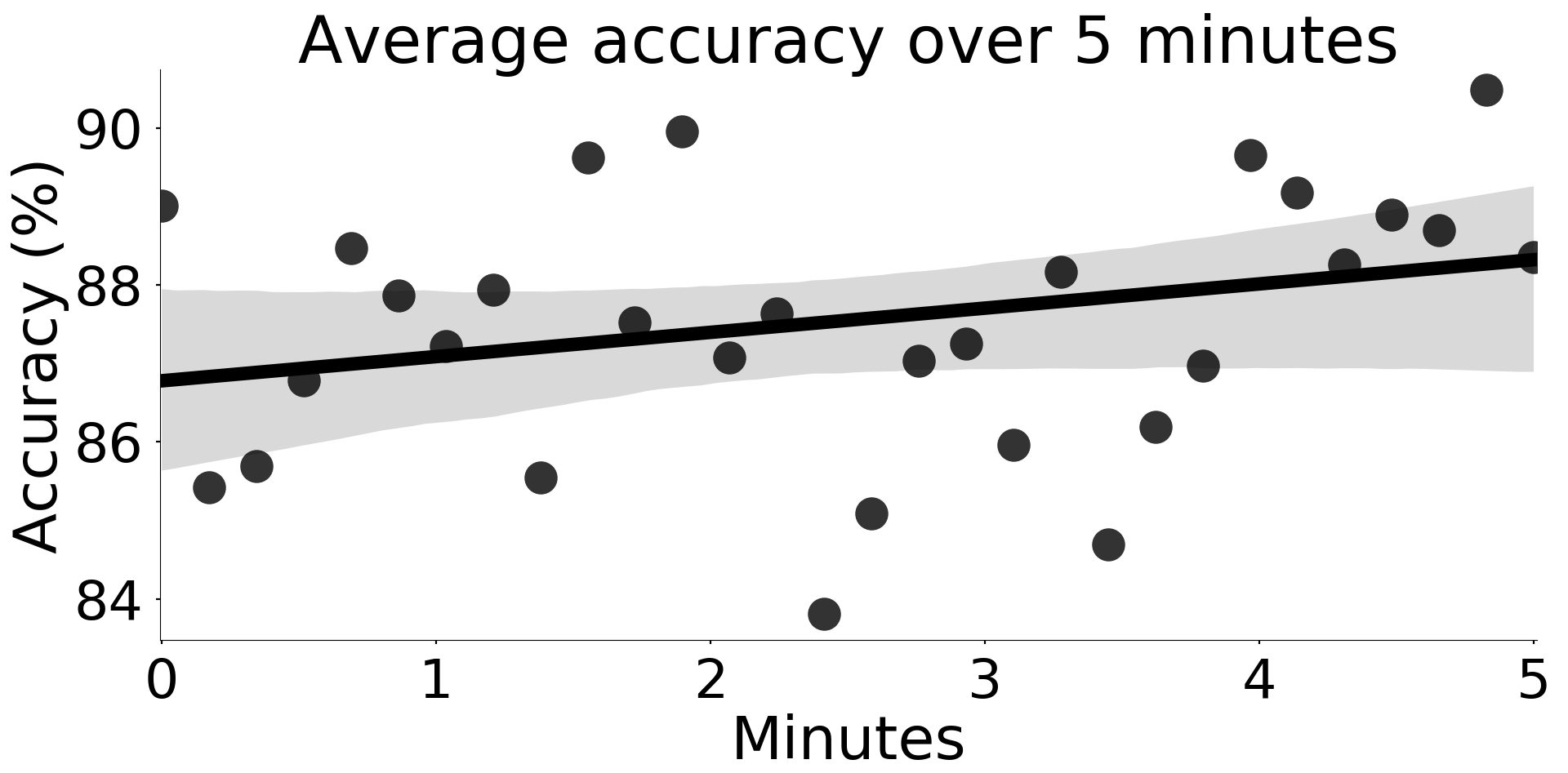}
    \caption{The average accuracy of the eight participants over all the five minute sessions recorded to evaluate the effect of muscle fatigue on the classifier performance.
    During each session of the experiment, participants were asked to hold a total of 30 random gestures for ten seconds each. As such, a dot represents the average accuracy across all participants over one of the ten second periods. The translucent bands around the linear regression represent the confidence intervals (95\%) estimated by bootstrap.}
    \label{five_minutes}
    \end{figure}

    Many participants reported experiencing muscular fatigue during the recording of both this experiment and the \textit{evaluation dataset}. As such, in an effort to quantify the impact of muscle fatigue on the classifier's performance, the average accuracy of the eight participants over the five minute session is computed as a function of time. As can be observed from the positive slope of the linear regression presented in Fig.~\ref{five_minutes}, muscle fatigue, does not seem to negatively affect the proposed ConvNet's accuracy. 
    
    \section{Discussion}
    \label{Discussion}
     Table~\ref{tableEvaluationDatasetTLComparison} and Table~\ref{NinaProTL_ConvNet_Table} show that, in all cases the TL augmented ConvNets significantly outperformed their non-augmented versions, regardless of the number of training cycles. As expected, reducing the amount of training cycles systematically degraded the performances of all tested methods (see Table~\ref{tableEvaluationDatasetTLComparison},~\ref{MyoDatasetComparisonStateOfTHeArtTable},~\ref{NinaProTL_ConvNet_Table},~\ref{NinaProComparisonStateOfTheArt} and Fig.~\ref{NinaProAccuracyGroup}), with the non-TL ConvNets being the most affected on the \textit{Myo Dataset}. This is likely due to overfitting that stems from the small size of the dataset. However, it is worth noting that, when using a single cycle of training, augmenting the ConvNets with the proposed TL scheme significantly improves their accuracies. In fact, with this addition, the accuracies of the ConvNets become the highest of all methods on both tested datasets. Overall, the proposed TL-augmented ConvNets were competitive with the current state-of-the-art, with the \textit{TL augmented CWT-based ConvNet} achieving a higher average accuracy than the traditional sEMG classification technique on both datasets for all training cycles. It is also noteworthy that while the \textit{raw+TL} ConvNet was the worst amongst the TL augmented ConvNet on the \textit{Myo Dataset}, it achieved the highest accuracy on the \textit{NinaPro DB5}. 
     Furthermore, the TL method outperformed the non-augmented ConvNets on the out-of-sample experiment. The difference in accuracy of the two methods was deemed significant by the Wilcoxon Signed Rank Test ($p<0.05$) for all training repetitions. This suggests that the proposed TL algorithm enables the network to learn features that can generalize not only across participants but also for never-seen-before gestures. As such, the weights learned from the \textit{pre-training dataset} can easily be re-used for other work that employs the Myo Armband with different gestures. 
     
     While in this paper, the proposed source and \textit{second network} were almost identical they are performing different tasks (see Sec.~\ref{proposed_TL_architecture}). As such further differentiation of both networks might lead to increased performance. At first glance, the element-wise summation between the \textit{source} and \textit{second network} might seem to impose a strong constraint on the architecture of the two networks. However, one could replace the learned scalar layers in the \textit{target network} by convolutions or fully connected layers to bridge the dimensionality gap between potentially vastly different \textit{source} and \textit{second} networks. 
     
     Additionally, a difference in the average accuracy between the real-time experiment (Sec.~\ref{medium_term_experiment}) and the \textit{Evaluation Dataset} (Sec.~\ref{comparison_deep_learning_MyoDataset}) was observed (95.42\% vs 98.31\% respectively). This is likely due to the reaction delay of the participants, but more importantly to the transition between gestures.
     These transitions are not part of the training dataset, because they are too time consuming to record as the number of possible transitions equals $n^2-n$ where $n$ is the number of gestures. Consequently, it is expected that the classifiers predictive power on transition data is poor in these circumstances. As such, being able to accurately detect such transitions in an unsupervised way might have a greater impact on the system's responsiveness than simply reducing the window size. This and the aforementioned point will be investigated in future works. 
     
     The main limitation of this study is the absence of tests with amputees. Additionally, the issue of electrode shifts has not been explicitly studied and the variability introduced by various limb positions was not considered when recording the dataset. A limitation of the proposed TL scheme is its difficulty to adapt when the new user cannot wear the same amount of electrodes as the group used for pre-training. This is because changing the number of channels changes the representation of the phenomena (i.e. muscle contraction) being fed to the algorithm. The most straightforward way of addressing this would be to numerically remove the relevant channels from the dataset used for pre-training. Then re-running the proposed TL algorithm on an architecture adapted to the new representation fed as input. Another solution is to consider the EMG channels in a similar way as color channels in image. This type of architecture seems, however, to perform worse than the ones presented in this paper (see Appendix~\ref{less_electrods_appendix}).
    
    \section{Conclusion}
    \label{conclusion}
    This paper presents three novel ConvNet architectures that were shown to be competitive with current sEMG-based classifiers. Moreover, this work presents a new TL scheme that systematically and significantly enhances the performances of the tested ConvNets. On the newly proposed \textit{evaluation} dataset, the TL augmented ConvNet achieves an average accuracy of 98.31\% over 17 participants. Furthermore, on the \textit{NinaPro DB5} dataset (18 hand/wrist gestures), the proposed classifier achieved an average accuracy of 68.98\% over 10 participants on a single Myo Armband. This dataset showed that the proposed TL algorithm learns sufficiently general features to significantly enhance the performance of ConvNets on out-of-sample gestures. Showing that deep learning algorithms can be efficiently trained, within the inherent constraints of sEMG-based hand gesture recognition, offers exciting new research avenues for this field.
    
    Future works will focus on adapting and testing the proposed TL algorithm on upper-extremity amputees. This will provide additional challenges due to the greater muscle variability across amputees and the decrease in classification accuracy compared to able-bodied participants~\cite{NinaPro_nature}. Additionally, tests for the application of the proposed TL algorithm for inter-session classification will be conducted as to be able to leverage labeled information for long-term classification. 
    

    \section{Acknowledgements}
    This research was supported by the Natural Sciences and Engineering Research Council of Canada (NSERC), [401220434], the Institut de recherche Robert-Sauv\'e en sant\'e et en s\'ecurit\'e du travail (IRSST), the Fondation Famille Choquette, and the Research Council of Norway through its Centres of Excellence scheme, project number 262762.
    
    \ifCLASSOPTIONcaptionsoff
      \newpage
    \fi

    \bibliographystyle{IEEEtran}
    \bibliography{main}

    
    \begin{appendices}
    
    \section{Data Augmentation}
    \label{data_augmentation}
    
    The idea behind data augmentation is to augment the size of the training set, with the objective of achieving better generalization.
    This is generally accomplished by adding realistic noise to the training data, which tends to induce a robustness to noise into the learned model.
    In many cases, this has been shown to lead to better generalization~\cite{data_augmentation_to_reduce_overfitting, dataAugmentation}. In this paper's context, data augmentation techniques can thus be viewed as part of the solution to reduce the overfitting from training a ConvNet on a small dataset. When adding noise to the data, it is important to ensure that the noise does not change the label of the examples.
    Hence, for image datasets, the most common and often successful techniques have relied on affine transformations~\cite{dataAugmentation}.
    
    Unfortunately, for sEMG signals, most of these techniques are unsuitable and cannot be applied directly. As such, specific data augmentation techniques must be employed. In this work, five data augmentation techniques are tested on the \textit{pre-training dataset} as they are part of the architecture building process. Note that this comparison was made with the ConvNet architecture presented in~\cite{smc_transfer_learning}, which takes as input a set of eight spectrograms (one for each channel of the Myo Armband).
    
    Examples are constructed by applying non-overlapping windows of 260\textit{ms}. This non-augmented dataset is referred to as the \textit{Baseline}. Consequently, an intuitive way of augmenting sEMG data is to apply overlapping windows (i.e. temporal translation) when building the examples. A major advantage of this technique within the context of sEMG signals - and time signals in general - is that it does not create any synthetic examples in the dataset compared to the affine transformation employed with images. Furthermore, with careful construction of the dataset, no new mislabeling occurs. In this work, this technique will be referred to as \textit{Sliding Window} augmentation. 
    
    Second, the effect of muscle fatigue on the frequency response of muscles fibers~\cite{muscle_fatigue} can be emulated, by altering the calculated spectrogram. The idea is to reduce the median frequency of a channel with a certain probability, by systematically redistributing part of the power of a frequency bin to an adjacent lower frequency one and so on. This was done in order to approximate the effect of muscle fatigue on the frequency response of muscle fibers~\cite{muscle_fatigue}.
    In this work, this technique will be referred to as \textit{Muscle Fatigue} augmentation.
    
    The third data augmentation technique employed aims at emulating electrode displacement on the skin.
    This is of particular interest, as the dataset was recorded with a dry electrode armband, for which this kind of noise is to be expected. The data augmentation technique consists of shifting part of the power spectrum magnitude from one channel to the next. In other words, part of the signal energy from each channel is sent to an adjacent channel emulating electrode displacement on the skin. 
    In this work, this approach will be referred to as \textit{Electrode Displacement} augmentation.
    
    For completeness, a fourth data augmentation technique which was proposed in a paper~\cite{CNN_NinaPro_60_percent} employing a ConvNet for sEMG gestures classification is also considered. The approach consists of adding a white Gaussian noise to the signal, with a signal-to-noise ratio of 25.
    This technique will be referred to as \textit{Gaussian Noise} augmentation. 
    
    Finally, the application of all these data augmentation methods simultaneously is referred to as the \textit{Aggregated Augmentation} technique.
    
    Data from these augmentation techniques will be generated from the \textit{pre-training dataset}. The data will be generated on the first two cycles, which will serve as the training set. The third cycle will be the validation set and the test set will be the fourth cycle. All augmentation techniques will generate double the amount of training examples compared to the baseline dataset. 
    
    Table~\ref{table_data_augmentation} reports the average test set accuracy for the 19 participants over 20 runs. In this appendix, the one-tail Wilcoxon signed rank test with Bonferroni correction is applied to compare the data augmentation methods with the baseline. The results of the statistical test are summarized in Table~\ref{table_data_augmentation}. The only techniques that produce significantly different results from the \textit{Baseline} is the \textit{Sliding Window} (improves accuracy). As such, as described in Sec.~\ref{dataset_recording_subsection} the only data augmentation technique employed in this work is the sliding windows.
    
    \begin{table*}[!htbp]
    \caption{Comparison of the five data augmentation techniques proposed.}
    \small
    \centering
    \begin{tabular}{@{}ccccccc@{}}
    \toprule
     & Baseline   & Gaussian Noise & Muscle Fatigue & Electrode Displacement & \multicolumn{1}{l}{Sliding Window} & Aggregated Augmentation \\ \midrule
    Accuracy                & 95.62\% & 93.33\%        & 95.75\%        & 95.80\%                & \textbf{96.14\%}                   & 95.37\%   \\
    STD           & 5.18\%  & 7.12\%         & 5.07\%         & 4.91\%                 & \textbf{4.93\%}                    & 5.27\%    \\
    Rank                         & 4       & 6              & 3              & 2                      & \textbf{1}                         & 5         \\$H_0$ (p-value) & -     & 1 
    & 1 
    & 1  
    & \multicolumn{1}{c}{\textbf{0
    (0.00542)}
    } & 1 
    \\ \bottomrule
    \end{tabular}
    \label{table_data_augmentation}

     The values reported are the average accuracies for the 19 participants over 20 runs.
     
     The Wilcoxon signed rank test is applied to compare the training of the ConvNet with and without one of the five data augmentation techniques. The null hypothesis is accepted when $H_0=1$ and rejected when $H_0=0$ (with $p=0.05$). As the \textit{Baseline} is employed to perform multiple comparison, Bonferroni correction is applied. As such, to obtain a global p-value of 0.05, a per-comparison p-value of 0.00833 is employed.
    
    \end{table*}
    \section{Deep Learning on Embedded Systems and real-time classification}
    \label{embeddedSystems}

    Within the context of sEMG-based gesture recognition, an important consideration is the feasibility of implementing the proposed ConvNets on embedded systems. As such, important efforts were deployed when designing the ConvNets architecture to ensure attainable implementation on currently available embedded systems. With the recent advent of deep learning, hardware systems particularly well suited for neural networks training/inference have been made commercially available. Graphics processing units (GPUs) such as the Nvidia Volta GV100 from \textit{Nvidia} (50 GFLOPs/s/W) \cite{nngpunvidia2017}, field programmable gate arrays (FPGAs) such as the Stratix 10 from \textit{Altera} (80 GFLOPs/s/W) \cite{nnfpga2017} and mobile system-on-chips (SoCs) such as the Nvidia Tegra from \textit{Nvidia} (100 GFLOPs/s/W) \cite{mobilesoc2015}, are commercially available platforms that target the need for portable, computationally efficient and low-power systems for deep learning inference. Additionally, dedicated Application-Specific Integrated Circuits (ASICs) have arisen from research projects capable of processing ConvNet orders of magnitudes bigger than the ones proposed in this paper at a throughput of 35 frames/s at 278\textit{mW}~\cite{eyeriss2017}. Pruning and quantizing network architectures are further ways to reduce the computational cost when performing inference with minimal impact on accuracy~\cite{pruning2016, quantizednn2016}.
    
    Efficient CWT implementation employing the Mexican Hat wavelet has already been explored for embedded platforms~\cite{optimizedcwt2016}. These implementations are able to compute the CWT of larger input sizes than those required in this work in less than 1\textit{ms}. Similarly, in~\cite{spectro_stftfpga2017}, a robust time-frequency distribution estimation suitable for fast and accurate spectrogram computation is proposed. To generate a classification, the proposed CNN-Spectrogram and CNN-CWT architectures (including the TL scheme proposed in Sec.~\ref{transfer_learning}) require approximately 14 728 000 and 2 274 000 floating point operations (FLOPs) respectively. Considering a 40\textit{ms} inference processing delay, hardware platforms of 3.5 and 0.5 GFLOPs/s/W will be suitable to implement a 100\textit{mW} embedded system for sEMG classification. As such, adopting hardware-implementation approaches, along with state-of-the-art network compression techniques will lead to a power-consumption lower than 100\textit{mW} for the proposed architectures, suitable for wearable applications.
    
    Note that currently, without optimization, it takes 21.42\textit{ms} to calculate the CWT and classify one example with the CWT-based ConvNet compared to 2.94\textit{ms} and 3.70\textit{ms} for the spectrogram and raw EMG Convnet respectively. Applying the proposed TL algorithm add an additional 0.57\textit{ms}, 0.90\textit{ms} and 0.14\textit{ms} to the computation for the CWT, spectrogram and raw EMG-based ConvNet respectively. These timing results were obtained by averaging the pre-processing and classifying time of the same 5309 examples across all methods. The gpu employed was a GeForce GTX 980M.
    
    \section{Feature Engineering}
    \label{feature_engineering}
    This section presents the features employed in this work. Features can be regrouped into different types, mainly: time, frequency and time-frequency domains. Unless specified otherwise, features are calculated by dividing the signal $x$ into overlapping windows of length $L$. The $kth$ element of the $ith$ window then corresponds to $x_{i,k}$.
    
    \subsection{Time Domain Features}
    
    \subsubsection{Mean Absolute Value (MAV)}~\cite{TD_stats}: A feature returning the mean of a fully-rectified signal.
    \begin{align}
            \text{MAV}(x_i) = \frac{1}{L}\sum_{k=1}^L |x_{i,k}|
    \end{align}
    
    \subsubsection{Slope Sign Changes (SSC)~\cite{TD_stats}}
    A feature that measures the frequency at which the sign of the signal slope changes. Given three consecutive samples $x_{i,k-1}$, $x_{i,k}$, $x_{i,k+1}$, the value of SSC is incremented by one if:
    \begin{align}
    (x_{i,k}-x_{i,k-1})*(x_{i,k}-x_{i,k+1}) \geq \epsilon
    \end{align}
    
    Where $\epsilon \geq 0$, is employed as a threshold to reduce the impact of noise on this feature. 
    
    \subsubsection{Zero Crossing (ZC)~\cite{TD_stats}}
    A feature that counts the frequency at which the signal passes through zero. A threshold $\epsilon \geq 0$ is utilized to lessen the impact of noise. The value of this feature is incremented by one whenever the following condition is satisfied: 
    
    \begin{align}
        (|x_{i,k} - x_{i, k+1}| \geq \epsilon) \wedge (sgn(x_{i,k}, x_{i,k+1}) \Leftrightarrow False)
    \end{align}
    
    Where $sgn(a, b)$ returns true if $a$ and $b$ (two real numbers) have the same sign and false otherwise. Note that depending on the slope of the signal and the selected $\epsilon$, the zero crossing point might not be detected. 
    
    \subsubsection{Waveform Length (WL)~\cite{TD_stats}} A feature that offers a simple characterization of the signal's waveform. It is calculated as follows:
    
    \begin{align}
        \text{WL}(x_i) = \sum_{k=1}^L |x_{i,k} - x_{i, k-1}|
    \end{align}
    
    \subsubsection{Skewness}
    The Skewness is the third central moment of a distribution which measures the overall asymmetry of a distribution. It is calculated as follows: 
    \begin{align}
        \text{Skewness}(x_i) = \frac{1}{L}\sum_{k=1}^{L}{\left(\frac{x_{i,k}-\overline{x}_{i}}{\sigma}\right)^3}
    \end{align}
    
    Where $\sigma$ is the standard deviation: 
    
    \subsubsection{Root Mean Square (RMS)~\cite{list_features}}
    This feature, also known as the quadratic mean, is closely related to the standard deviation as both are equal when the mean of the signal is zero. RMS is calculated as follows: 
    \begin{align}
        \text{RMS}(x_i) = \sqrt{\frac{1}{L}\sum_{k=1}^{L}x^2_{i,k}}
    \end{align}
    
    \subsubsection{Hjorth Parameters~\cite{hjorth_parameters}} Hjorth parameters are a set of three features originally developed for characterizing electroencephalography signals and then successfully applied to sEMG signal recognition~\cite{first_emg_hjorth, article_8}. \textit{Hjorth Activity Parameter} can be thought of as the surface of the power spectrum in the frequency domain and corresponds to the variance of the signal calculated as follows:
    
    \begin{align}
        \text{Activity}(x_i) = \frac{1}{L}\sum_{k=1}^{L}{(x_{i,k}-\overline{x}_{i})^2}
    \end{align}
    Where $\overline{x}_{i}$ is the mean of the signal for the $ith$ window. \textit{Hjorth Mobility Parameter} is a representation of the mean frequency of the signal and is calculated as follows: 
    
    \begin{align}
    \text{Mobility}(x_i) = \sqrt{\frac{\text{Activity}(x^{'}_{i})}{\text{Activity}(x_i)}}
    \end{align}
    
    Where $x^{'}_{i}$ is the first derivative in respect to time of the signal for the $ith$ window. Similarly, the \textit{Hjorth Complexity Parameter}, which represents the change in frequency, is calculated as follows:
    
    \begin{align}
        \text{Complexity}(x_i) = \frac{\text{Mobility}(x^{'}_{i})}{\text{Mobility}(x_i)}
    \end{align}
    
    \subsubsection{Integrated EMG (IEMG)}~\cite{list_features}:
    A feature returning the sum of the fully-rectified signal.
    \begin{align}
        \text{IEMG}(x_i) = \sum_{k=1}^L |x_{i,k}|
    \end{align}
    
    \subsubsection{Autoregression Coefficient (AR)}~\cite{complex_features_explanation}
    An autoregressive model tries to predict future data, based on a weighted average of the previous data. This model characterizes each sample of the signal as a linear combination of the previous sample with an added white noise. The number of coefficients calculated is a trade-off between computational complexity and predictive power. The model is defined as follows: 
    
    \begin{align}
        x_{i,k} = \sum_{j=1}^{P} \rho_j  x_{i,k-j} + \epsilon_t
    \end{align}
    
    Where P is the model order, $\rho_j$ is the $jth$ coefficient of the model and $\epsilon_t$ is the residual white noise.

    \subsubsection{ Sample Entropy (SampEn)}~\cite{sampen_description}
    Entropy measures the complexity and randomness of a system. Sample Entropy is a method which allows entropy estimation. 
    
    \begin{align}
        \text{SampEn}(x_{i}, m, r) = -\ln\left(\frac{A^m(r)}{B^m(r)}\right)
    \end{align}
    
    \subsubsection{EMG Histogram (HIST)~\cite{emg_histogram}}
    
    When a muscle is in contraction, the EMG signal deviates from its baseline. The idea behind HIST is to quantify the frequency at which this deviation occurs for different amplitude levels. HIST is calculated by determining a symmetric amplitude range centered around the baseline. This range is then separated into $n$ bins of equal length ($n$ is a hyperparameter). The HIST is obtained by counting how often the amplitude of the signal falls within each bin's boundaries.

    \subsection{Frequency Domain Features}
    
    \subsubsection{Cepstral Coefficient~\cite{ar_cepstral, complex_features_explanation}}
    
    The cepstrum of a signal is the inverse Fourier transform of the log power spectrum magnitude of the signal. Like the AR, the coefficients of the cepstral coefficients are employed as features. They can be directly derived from AR as follows:
    
    \begin{align}
            c_1 = -a_1
    \end{align}
    \begin{align}
            c_i = -a_i - \sum_{n=1}^{i-1}(1 - \frac{n}{i})a_n c_{i-n} \text{     ,with 1 $<$ i$\leq$ P}
    \end{align}
    
    \subsubsection{Marginal Discrete Wavelet Transform (mDWT)~\cite{marginalDWT}}
    
    The mDWT is a feature that removes the time-information from the discrete wavelet transform to be insensitive to wavelet time instants. The feature instead calculates the cumulative energy of each level of the decomposition. The computation of the mDWT for each channel is implemented as follow in~\cite{NinaProDB5} (See Algorithm \ref{mDWT_algo}).
    
    \begin{algorithm}
    \caption{mDWT pseudo-code}
    \label{mDWT_algo}
    \begin{algorithmic}[1]
    \Procedure{mDWT}{}
        \State{$wav \gets db7$}
        \State{$level \gets 3$}
        \State{$coefficients \gets wavDec(x, level, wav)$}
        \State{$N \gets length(coefficients)$}
        \State{$SMax \gets log2(N)$}
        \State{$Mxk \gets []$}
        \For {s=1,...,SMax}
            \State{$CMax \gets \frac{N}{2^S}-1$}
            \State{$val \gets \sum_{u=0}^{CMax} |coefficients[u]|$}
            \State{$Mxk.append(val)$}
        \EndFor
        \Return $Mxk$
    \EndProcedure
    \end{algorithmic}
    \end{algorithm}
    
    Where $x$ is the 1-d signal from which to calculate the mDWT and wavDec is a function that calculates the wavelet decomposition of a vector at level $n$ using the wavelet $wav$. The coefficients are returned in a 
    1-d vector with the \textit{Approximation Coefficients}($AC$) placed first followed by the \textit{Detail Coefficients}($DC$) (i.e. $coefficients = [CA, CD3, CD2, CD1]$, where 3, 2, 1 are the level of decomposition of the $DC$).
    
    Note that due to the choice of the level (3) of the wavelet decomposition in conjunction with the length of $x$ (52) in this paper, the mDWT will be affected by boundaries effects. This choice was made to be as close as possible to the mDWT features calculated in~\cite{NinaProDB5} which employed the same wavelet and level on a smaller $x$ length (40).

    
    
    \subsection{Time-Frequency Domain Features}
    
    \subsubsection{Short Term Fourier Transform based Spectrogram (Spectrogram)}
    The Fourier transform allows for a frequency-based analysis of the signal as opposed to a time-based analysis. However, by its nature, this technique cannot detect if a signal is non-stationary. As sEMG are non-stationary~\cite{complexity_EMG}, an analysis of these signals employing the Fourier transform is of limited use. An intuitive technique to address this problem is the STFT, which consists of separating the signal into smaller segments by applying a sliding window where the Fourier transform is computed for each segment. In this context, a window is a function utilized to reduce frequency leakage and delimits the segment's width (i.e. zero-valued outside of the specified segment). The spectrogram is calculated by computing the squared magnitude of the STFT of the signal. In other words, given a signal $s(t)$ and a window of width $w$, the spectrogram is then:
    \begin{align}
    spectrogram(s(t), w) = |STFT(s(t),w)|^2
    \end{align}

    \subsubsection{Continuous Wavelet Transform (CWT)}
    The Gabor limit states that a high resolution both in the frequency and time-domain cannot be achieved~\cite{gabor_limit}. Thus, for the STFT, choosing a wider window yields better frequency resolution to the detriment of time resolution for all frequencies and vice versa. 
    
    Depending on the frequency, the relevance of the different signal's attributes change. Low-frequency signals have to be well resolved in the frequency band, as signals a few~\textit{Hz} apart can have dramatically different origins (e.g. Theta brain waves (4 to 8\textit{Hz}) and Alpha brain waves (8 to 13\textit{Hz})~\cite{brain_waves}). On the other hand, for high-frequency signals, the relative difference between a few or hundreds~\textit{Hz} is often irrelevant compared to its resolution in time for the characterization of a phenomenon.
    
    \begin{figure}[!htbp]
    \centering
    \includegraphics[width=.8\linewidth]{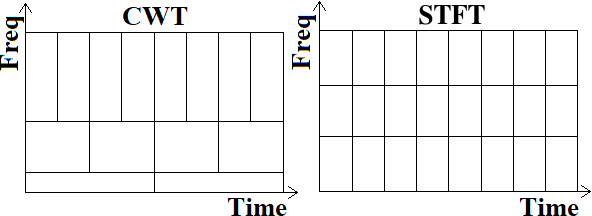}
    \caption{A visual comparison between the CWT and the STFT. Note that due to its nature, the \textit{frequency} of the CWT is, in fact, a pseudo-frequency.}
    \label{cwt_vs_stft}
    \end{figure}
    As illustrated in Fig.~\ref{cwt_vs_stft}, this behavior can be obtained by employing \textit{wavelets}. A wavelet is a signal with a limited duration, varying frequency and a mean of zero~\cite{processingBible}. The \textit{mother wavelet} is an arbitrarily defined wavelet that is utilized to generate different wavelets. The idea behind the wavelets transform is to analyze a signal at different scales of the mother wavelet~\cite{waveletBible}. For this, a set of wavelet functions are generated from the mother wavelet (by applying different scaling and shifting on the time-axis). The CWT is then computed by calculating the convolution between the input signal and the generated wavelets.
    
    \section{Hyperparameters selection for state of the art feature sets.}
    \label{hyperparameters_selection}
     The hyperparameters considered for each classifiers were as follow: 
    \begin{itemize}
        \item SVM: Both the RBF and Linear kernel were considered. The soft margin tolerance ($C$) was chosen between $10^{-3}$ to $10^3$ on a logarithm scale with 20 values equally distributed. Similarly the $\gamma$ hyperparameter for the RBF kernel was selected between $10^{-5}$ to $10^2$ on a logarithm scale with 20 values equally distributed.
        \item ANN: The size of the hidden layers was selected between $20$ to $1500$ on a logarithm scale with 20 values equally distributed. The activation functions considered were sigmoid, tanh and relu. The learning rate was initialized between $10^{-4}$ to $10^0$. The L2 penalty was selected between $10^{-6}$ to $10^{-2}$ with 20 values. Finally, the solver employed is Adam and early stopping is applied using 10\% of the training data as validation. 
        \item KNN: The number of possible neighbors considered were 1, 2, 3, 4, 5, 10, 15 and 20. The metric distance considered was the Manhattan distance, the euclidean distance and the Minkowski distance of the third and fourth degree. 
        \item RF: The range of estimators considered were between 5 to 1000 using a logarithm scale with 100 values equally distributed. The maximum number of features considered (expressed as a ratio of the total number of features fed to the RF) were: .1, .2, .3, .4, .5, .6, .7, .8, .9, 1. Additionally, both the square root and the $log_2$ of the total number of features fed to the RF were also considered.
        
    \end{itemize}
    Note that the hyperparameter ranges for each classifier were chosen using 3 fold cross-validation on the \textit{pre-training dataset}. 
    
    \section{Dimensionality Reduction on the Myo Armband Dataset for State of the Art Feature Set}
    \label{Dimensionality_reduction}
    Table~\ref{dim_reduction_table} shows the average accuracies obtained on the \textit{Evaluation dataset} for the state-of-the-art feature sets with and without dimensionality reduction. Note that all the results with dimensionality reduction were obtained in a week of computation. In contrast, removing the dimensionality reduction significantly augmented the required time to complete the experiments to more than two and a half months of continuous run time on an AMD-Threadripper 1900X 3.8Hz 8-core CPU.

    \begin{table*}[!htbpt!]
    \centering
    \caption{Classification accuracy on the \textit{Evaluation dataset} for the feature sets with and without dimensionality reduction.}
    \begin{tabular}{@{}ccccccccc@{}}
    \toprule
    \multicolumn{1}{c|}{} & \multicolumn{2}{c|}{TD} & \multicolumn{2}{c|}{Enhanced TD} & \multicolumn{2}{c|}{Nina Pro} & \multicolumn{2}{c}{SampEn Pipeline} \\ \midrule
     & \begin{tabular}[c]{@{}c@{}}With \\ Dimensionality\\  Reduction\end{tabular} & \begin{tabular}[c]{@{}c@{}}Without\\ Dimensionality\\  Reduction\end{tabular} & \begin{tabular}[c]{@{}c@{}}With \\ Dimensionality\\  Reduction\end{tabular} & \begin{tabular}[c]{@{}c@{}}Without\\ Dimensionality\\  Reduction\end{tabular} & \begin{tabular}[c]{@{}c@{}}With \\ Dimensionality\\  Reduction\end{tabular} & \begin{tabular}[c]{@{}c@{}}Without\\ Dimensionality\\  Reduction\end{tabular} & \begin{tabular}[c]{@{}c@{}}With \\ Dimensionality\\  Reduction\end{tabular} & \begin{tabular}[c]{@{}c@{}}Without\\ Dimensionality\\  Reduction\end{tabular} \\
    4 Cycles & \begin{tabular}[c]{@{}c@{}}\textbf{97.76\%}\\ \textbf{(LDA)}\end{tabular} & \begin{tabular}[c]{@{}c@{}}96.74\%\\ (KNN)\end{tabular} & \begin{tabular}[c]{@{}c@{}}\textbf{98.14\%}\\ \textbf{(LDA)}\end{tabular} & \begin{tabular}[c]{@{}c@{}}96.85\%\\ (RF)\end{tabular} & \begin{tabular}[c]{@{}c@{}}\textbf{97.58\%}\\ \textbf{(LDA)}\end{tabular} & \begin{tabular}[c]{@{}c@{}}97.14\%\\ (RF)\end{tabular} & \begin{tabular}[c]{@{}c@{}}\textbf{97.72\%}\\ \textbf{(LDA)}\end{tabular} & \begin{tabular}[c]{@{}c@{}}96.72\%\\ (KNN)\end{tabular} \\
    3 Cycles & \begin{tabular}[c]{@{}c@{}}\textbf{96.26\%}\\ \textbf{(KNN)}\end{tabular} & \begin{tabular}[c]{@{}c@{}}96.07\%\\ (RF)\end{tabular} & \begin{tabular}[c]{@{}c@{}}\textbf{97.33\%}\\ \textbf{(LDA)}\end{tabular} & \begin{tabular}[c]{@{}c@{}}95.78\%\\ (RF)\end{tabular} & \begin{tabular}[c]{@{}c@{}}\textbf{96.54\%}\\ \textbf{(KNN)}\end{tabular} & \begin{tabular}[c]{@{}c@{}}96.53\%\\ (RF)\end{tabular} & \begin{tabular}[c]{@{}c@{}}\textbf{96.51\%}\\ \textbf{(KNN)}\end{tabular} & \begin{tabular}[c]{@{}c@{}}95.90\%\\ (KNN)\end{tabular} \\
    2 Cycles & \begin{tabular}[c]{@{}c@{}}\textbf{94.12\%}\\ \textbf{(KNN)}\end{tabular} & \begin{tabular}[c]{@{}c@{}}93.45\%\\ (RF)\end{tabular} & \begin{tabular}[c]{@{}c@{}}\textbf{94.79\%}\\ \textbf{(LDA)}\end{tabular} & \begin{tabular}[c]{@{}c@{}}93.06\%\\ (RF)\end{tabular} & \begin{tabular}[c]{@{}c@{}}93.82\%\\ (KNN)\end{tabular} & \begin{tabular}[c]{@{}c@{}}\textbf{94.25\%}\\ \textbf{(SVM)}\end{tabular} & \begin{tabular}[c]{@{}c@{}}\textbf{94.64\%}\\ \textbf{(KNN)}\end{tabular} & \begin{tabular}[c]{@{}c@{}}93.23\%\\ (KNN)\end{tabular} \\
    1 Cycle & \begin{tabular}[c]{@{}c@{}}\textbf{90.62\%}\\ \textbf{(KNN)}\end{tabular} & \begin{tabular}[c]{@{}c@{}}89.28\%\\ (KNN)\end{tabular} & \begin{tabular}[c]{@{}c@{}}\textbf{91.25\%}\\ \textbf{(LDA)}\end{tabular} & \begin{tabular}[c]{@{}c@{}}88.63\%\\ (SVM)\end{tabular} & \begin{tabular}[c]{@{}c@{}}90.13\%\\ (LDA)\end{tabular} & \begin{tabular}[c]{@{}c@{}}\textbf{90.32\%}\\ \textbf{(SVM)}\end{tabular} & \begin{tabular}[c]{@{}c@{}}\textbf{91.08\%}\\ \textbf{(KNN)}\end{tabular} & \begin{tabular}[c]{@{}c@{}}89.27\%\\ (KNN)\end{tabular} \\ \bottomrule
    \end{tabular}
    \label{dim_reduction_table}
    \end{table*}
    
    \section{Reducing the number of EMG channels on the target dataset}
    \label{less_electrods_appendix}
    
    If the new user cannot wear the same amount of electrodes as what was worn during pre-training the proposed transfer learning technique cannot be employed out of the box. A possible solution is to consider that the EMG channels are akin to the channel of an image, giving different view of the same phenomenon. In this section, the enhanced raw ConvNet is modified to accommodate this new representation. The 2D image (8 x 52) that was fed to the network is now a 1D image (of length 52) with 8 channels. The architecture now only employs 1D convolutions (with the same parameters). Furthermore, the amount of neurons in the fully connected layer was reduced from 500 to 256. The \textit{second network} is identical to the \textit{source network}. 
    
    Pre-training is done on the \textit{pre-training dataset}, training on the first \textit{round} of the \textit{evaluation dataset} with 4 cycles of training and the test is done on the last two \textit{rounds} of the \textit{evaluation dataset}. The first, third, fifth and
    eighth channels are removed from every participant on the \textit{evaluation dataset}. The \textit{pre-training dataset} remains unchanged.
    
    The non-augmented ConvNet achieves an average accuracy of 61.47\% over the 17 participants. In comparison, the same network enhanced by the proposed transfer learning algorithm achieves an average accuracy of 67.65\% accuracy. This difference is judged significant by the one-tail Wilcoxon Signed Rank Test (p-value=0.00494). While the performance of this modified ConvNet is noticeably lower than the other classification methods viewed so far it does show that the proposed TL algorithm can be adapted to different numbers of electrodes between the source and the target.
    \end{appendices}
    
    \end{document}